\documentclass[letterpaper]{article} 
\usepackage{aaai25}  
\usepackage{times}  
\usepackage{helvet}  
\usepackage{courier}  
\usepackage[hyphens]{url}  
\usepackage{graphicx} 
\usepackage{natbib}  
\usepackage{caption} 
\usepackage{algorithm}
\usepackage{algorithmic}
\usepackage{amsmath}
\usepackage{tabularx}
\usepackage{multirow}
\usepackage{multicol}
\usepackage{booktabs}
\usepackage{collcell}
\usepackage[table]{xcolor}
\usepackage{xspace} 
\usepackage{subfigure}
\usepackage{amsfonts}

\usepackage{newfloat}
\usepackage{listings}
\DeclareCaptionStyle{ruled}{labelfont=normalfont,labelsep=colon,strut=off} 
\floatstyle{ruled}
\newfloat{listing}{tb}{lst}{}
\floatname{listing}{Listing}

\title{Make Domain Shift a Catastrophic Forgetting Alleviator in Class-Incremental Learning}
\author {
    Wei Chen\textsuperscript{\rm 1,2},
    Yi Zhou\textsuperscript{\rm 1,2 \thanks{Corresponding author}}
}
\affiliations {
    \textsuperscript{\rm 1}School of Computer Science and Engineering, Southeast University, China\\
    \textsuperscript{\rm 2}Key Laboratory of New Generation Artificial Intelligence Technology and Its Interdisciplinary Applications, Ministry of Education, China\\
    weighchen@seu.edu.cn, yizhou.szcn@gmail.com
}

\begin{document}
\frenchspacing
\newcommand{\methodname}{DisCo\xspace}
\newcommand{\gray}[1]{\cellcolor{gray!30}#1}
\maketitle

\begin{abstract}
In the realm of class-incremental learning (CIL), alleviating the catastrophic forgetting problem is a pivotal challenge. This paper discovers a counter-intuitive observation: by incorporating domain shift into CIL tasks, the forgetting rate is significantly reduced. Our comprehensive studies demonstrate that incorporating domain shift leads to a clearer separation in the feature distribution across tasks and helps reduce parameter interference during the learning process. Inspired by this observation, we propose a simple yet effective method named \methodname to deal with CIL tasks. \methodname introduces a lightweight prototype pool that utilizes contrastive learning to promote distinct feature distributions for the current task relative to previous ones, effectively mitigating interference across tasks. \methodname can be easily integrated into existing state-of-the-art class-incremental learning methods. Experimental results show that incorporating our method into various CIL methods achieves substantial performance improvements, validating the benefits of our approach in enhancing class-incremental learning by separating feature representation and reducing interference. These findings illustrate that \methodname can serve as a robust fashion for future research in class-incremental learning.
\end{abstract}
\begin{links}
    \link{Code}{https://github.com/PixelChen24/DisCo}
\end{links}

\section{Introduction}
Deep neural networks excel in static environments but falter with the dynamic nature of real-world data. Designed to learn from static datasets, these models struggle to adapt to new data without complete retraining. This leads to performance degradation and catastrophic forgetting \cite{catastrophic1} in real-world applications with continuously updated data~\cite{survey_ad}. Continual learning, also known as incremental or lifelong learning, addresses this by allowing models to learn incrementally, retaining knowledge over time, and adapting to evolving data.

Generally, continual learning can be mainly taxonomized as \textbf{C}lass-\textbf{I}ncremental \textbf{L}earning(CIL) and \textbf{D}omain-\textbf{I}ncremental \textbf{L}earning(DIL). CIL~\cite{memo, ewc, architecture3, L2P} involves learning a sequence of tasks, with each task only introducing new classes that were not present in the previous tasks. The model should correctly classify new samples into all classes seen so far. Unlike CIL which focuses on expanding the model's class knowledge, DIL~\cite{DIL1, DIL2, DIL3} requires the model to generalize effectively across varying data distributions~\cite{lai_domainshift, liu_domainshift} with the same label space and fight against forgetting at the same time.
\begin{figure}[t]
  \centering
  \includegraphics[width=1.0\columnwidth]{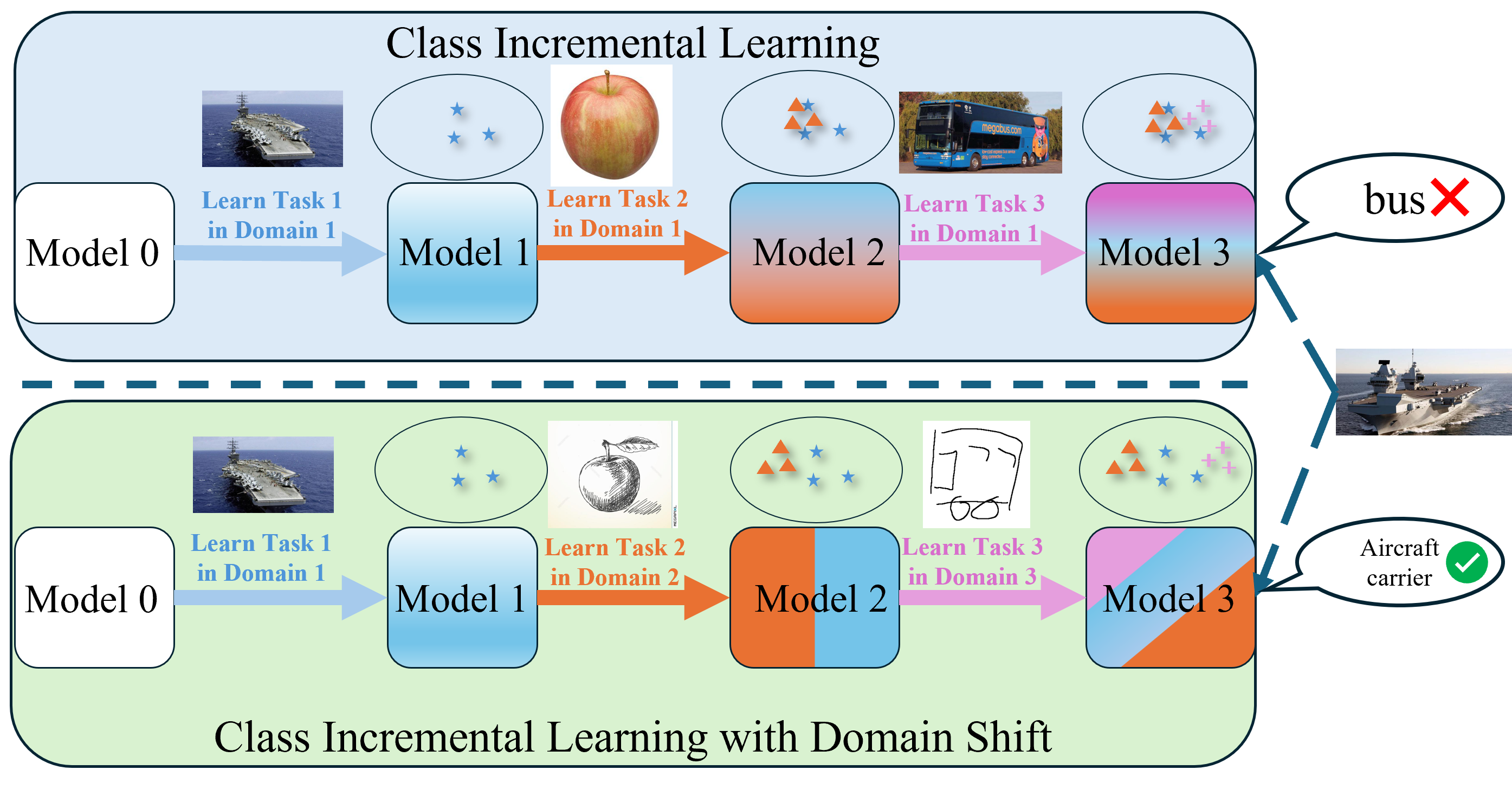}
    \caption{The key finding of our work: Incorporating domain shift in class incremental learning contributes to a clear separation of feature space and a better resistance to forgetting.}
    \label{fig: introduction_illustration}
\end{figure}
There is no doubt that combination of CIL and DIL can better simulate the real world where new knowledge and new data distribution are introduced gradually. Relatively a few studies have been conducted to provide solutions to this combination. Many works~\cite{UCL1, UCL2, UCL3} explore various problem settings of combining CIL with DIL, emphasizing adaptation and generalization across known and unknown domains. However, they did not really focus on the \textbf{effect of domain shift on CIL}.  Thus, we are motivated by this question: ``\textit{Does domain shift really hamper CIL methods?}''

\label{sec: introduction}

In this work, we first discover that domain shift helps reduce the interference and forgetting across tasks in CIL, as illustrated in Fig. ~\ref{fig: introduction_illustration}. Specifically, we start with an empirical study where we simulate the combination of CIL and domain shift by splitting DomainNet~\cite{domainnet} dataset, or manually add domain shift using a style transfer GAN to the original CIFAR-100 dataset~\cite{cifar100} to construct DomainCIFAR-100. Qualitative results on DomainNet and DomainCIFAR-100 show that domain shift assists model in learning distinguishable representations across tasks. To further investigate the effects of domain shift, we design a quantitative metric to measure the interference across tasks, along with a metric to measure the knowledge learned by the model. We find that domain shift helps reduce the interference during parameter updating and improves the knowledge transfer, thus reducing forgetting rate. Then, we leverage this discovery to design a simple yet effective plug-and-play method named \textbf{DisCo} (\textbf{Dis}tinguishable feature for \textbf{Co}ntinual Learning) to deal with CIL tasks. \methodname utilizes contrastive loss to impose task-level and class-level regularization on prototypes to foster distinguishable task representations. \methodname also includes a cross-task contrastive distillation loss to preserve prior knowledge effectively. \methodname can be easily integrated into existing state-of-the-art continual learning methods, especially rehearsal-based methods, to boost performance. We perform extensive experiments on popular CIL benchmarks and show that incorporating \methodname reduces forgetting significantly and improves their performance consistently. 

\textbf{Our contributions are highlighted as follows:}
\textbf{1)} We observe the counter-intuitive phenomenon that when introducing domain shift to the standard CIL setting, the overall forgetting is significantly reduced. To the best of our knowledge, we're the first to discover this phenomenon.
\textbf{2)} Through analyzing the parameter updated during sequential tasks, we find that the task interference is small due to the variance caused by the domain shift of input, which consequently leads to a relatively lower forgetting rate.
\textbf{3)} Based on our observation, we introduce a simple yet effective plug-and-play method named \methodname, which can be easily integrated into existing class-incremental methods to hedge against forgetting.

\section{Related Works}
\textbf{Class-Incremental Learning.}
\label{sec: related works: cil}
In class-incremental learning, models are continuously updated with new class data, aiming to retain performance on previously learned classes without the original training data. Various strategies address forgetting~\cite{zhangxingxing}, including rehearsal-based methods, which use a memory buffer to store exemplars~\cite{icarl, AQM, memo} or generate images of old classes using generative networks \cite{generative2, generative1}. Regularization-based approaches implement weight regularization on important parameters~\cite{ ewc, regularization-weight} or knowledge distillation to preserve crucial outputs \cite{lwf,regularization-function}. Architecture-based methods expand\cite{DER, architecture2, architecture3} or reallocate\cite{architecture1} the model's structure to accommodate new tasks. Meanwhile, recently popular prompt-based methods~\cite{codaprompt, dualprompt, progressiveprompt}, such as L2P~\cite{L2P} guide pre-trained Transformers with task-specific prompts to balance shared and task-specific knowledge. Mixed strategies, like DER\cite{DER}, combine two or more strategies above to achieve a more robust continual learner.

 \textbf{Contrastive Learning in Continual Learning.} Contrastive loss has been integrated into continual learning methods to combat catastrophic forgetting, with approaches like Co2L \cite{co2l}, which utilizes supervised contrastive loss for task learning paired with self-supervised loss for knowledge distillation between models. DualNet \cite{dualnet} employs both supervised and self-supervised losses in training its fast and slow learners respectively, enhancing generalizable representations. These methods operate on the assumption that contrastive loss yields more stable representations for future tasks compared to cross-entropy loss \cite{co2l}. Our research, however, focuses on the utility of contrastive loss in learning discriminative features across tasks.

\textbf{Domain Shift in Class-Incremental Learning.} The intersection of class-incremental and domain-incremental learning has been sparingly explored. Kundu's work~\cite{UCL1} blends class-incremental learning with source-target domain adaptation, specifically designed for open-set environments. Meanwhile, Xie~\cite{UCL3} has crafted a comprehensive framework that concurrently addresses the challenges posed by both class and domain continual learning. Building on this, Simon~\cite{UCL2} introduces a method that not only addresses cross-domain continual learning but also ensures robust generalization to new, unseen domains. Despite these innovative approaches, the literature still lacks a detailed exploration of how domain shift specifically affects class-incremental learning.

\section{Empirical Study}

\subsection{Problem Setup}
Here we first introduce the formal definition of standard \textbf{C}lass-\textbf{I}ncremental \textbf{L}earning (\textbf{CIL}). Let \( \mathcal{D} = \{D_1, D_2, ..., D_{T}\} \) represent a sequence of datasets corresponding to tasks \( 1,2,...T\). Each dataset \( D_t = \{(x_i^t, y_i^t)\}_{i=1}^{N_t} \) consists of \( N_t \) samples, where \( x_i^t \) is the \( i \)-th input and \( y_i^t \) is the corresponding label from the label set \( \mathcal{C}_t\). For each task \( T_t \), the label space \( \mathcal{C}_t \) introduces new classes, and \( \mathcal{C}_t \cap \mathcal{C}_{t'} = \emptyset \) for \( t \neq t' \). Thus, the cumulative label space up to task \( t \) is \( \mathcal{C}^t = \bigcup_{k=1}^t \mathcal{C}_k \). Model at task $t$ only has access to $D_t$, and the goal at task $t$ is to train a model \( f_{\theta}^t \)\label{sec: Problem setup} parameterized by \( \theta_t \) which can classify inputs \( x \) into the correct class among all classes \( \mathcal{C}^t \) seen so far. Formally, after training on \( D_T \), the model \( f_\theta ^T \) should minimize the loss:
{
\begin{align}
     \mathcal{L}(\theta) = \sum_{t=1}^{T} \sum_{(x_i^t, y_i^t) \in D_t} L(f_\theta^T(x_i^t), y_i^t),
\end{align}
}
where \( L \) is a loss function appropriate for classification.

Most previous CIL works conduct experiments under the setting that all tasks share the same distribution, i.e. \(\mathcal{P}_t=\mathcal{P}_{t'}\), for \(t\neq t'\) and overlook the effect of domain shift on CIL. We are interested in this question: What if the \(\mathcal{P}\) is different from each other? i.e., \(\mathcal{P}_t\neq\mathcal{P}_{t'},\  \forall t\neq t'\).

\begin{figure*}[ht]
    \centering
    \subfigure[CIL and CILD scenario on CIFAR-100.]{
        \centering
        \includegraphics[width=1.0\columnwidth]{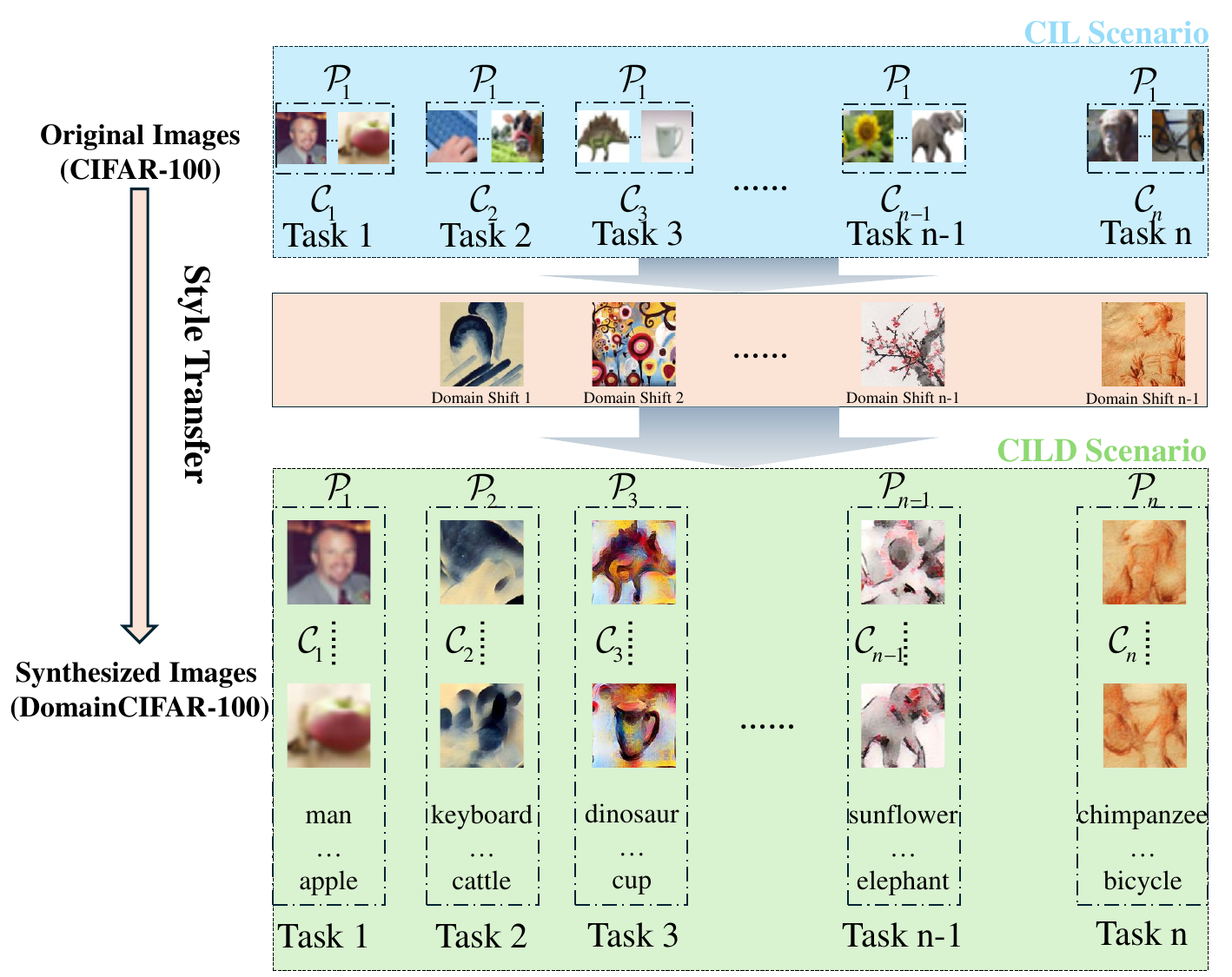}
        \label{subfig: cifar100}
    }
    \subfigure[CIL and CILD scenario on DomainNet.]{
        \centering
        \includegraphics[width=1.0\columnwidth]{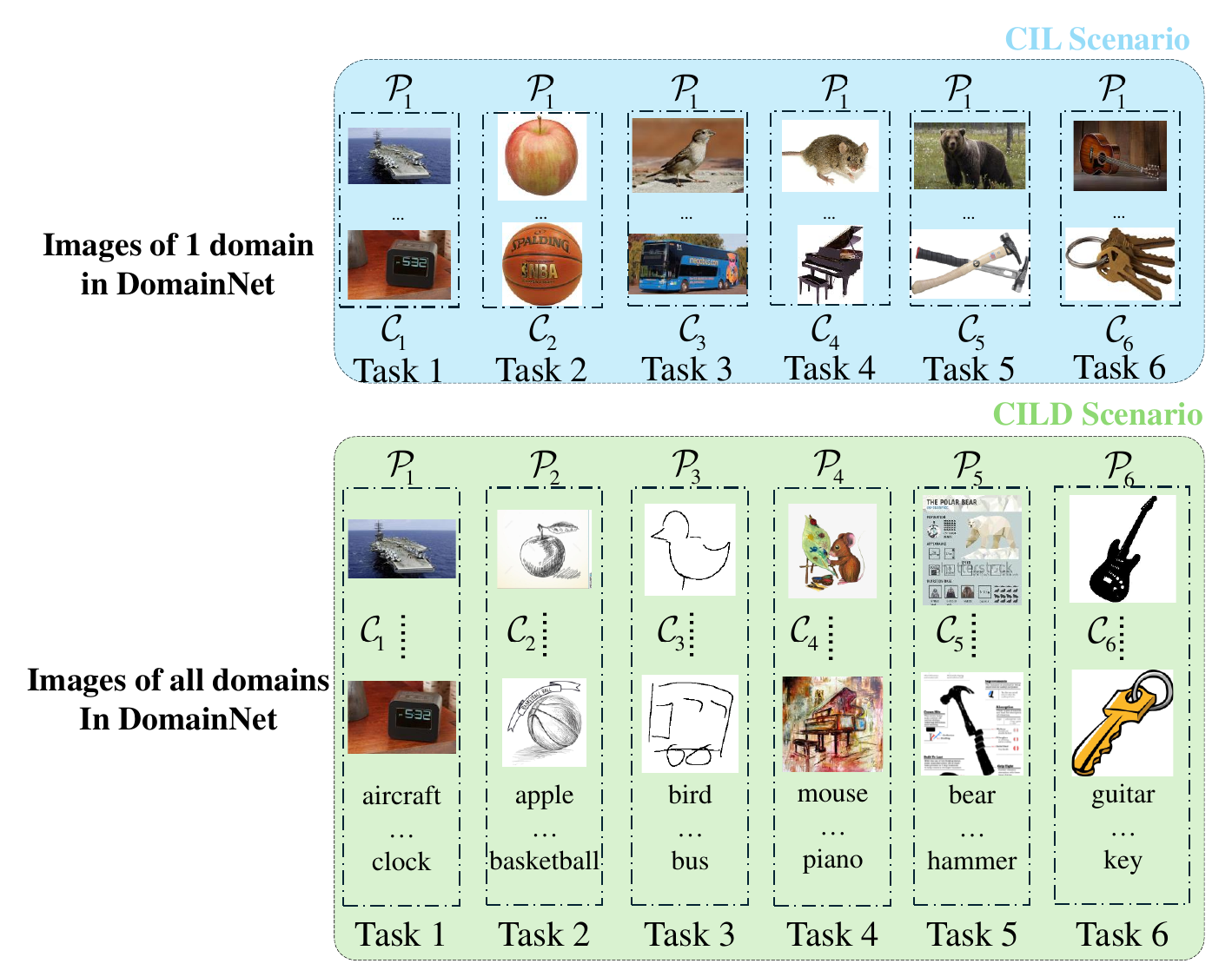}
        \label{subfig: domainnet}
    }
    \caption{
        Illustration of two scenarios construction on two datasets respectively. In Fig. ~\ref{subfig: cifar100}, we use AvatarNet~\cite{avatarnet} to synthesize images of new domains on CIFAR-100. We use the original images as training/testing set for CIL scenario and synthesized images(termed DomainCIFAR-100) as training/testing set for CILD scenario. In Fig. ~\ref{subfig: domainnet}, we split DomainNet~\cite{domainnet} to construct CIL and CILD scenario. Images of one domain make up the training/testing set for CIL scenario, and images of all domains make up the training/testing set for CILD scenario. In each dataset, the label space $\mathcal{C}_t$ of each task $t$ in CILD is consistent with that of CIL.
    }
    \label{fig: Illustration of CILD}

\end{figure*}

\subsection{Observation of Domain Shift on CIL}
\label{sec: empirical_study}
To investigate the effect of domain shift in class-incremental learning methods, we first illustrate some analytical experiments on CIFAR-100~\cite{cifar100} and DomainNet~\cite{domainnet} under two scenarios.
\subsubsection{Empirical Study Setup.}
We design two comparative scenarios: the classic CIL scenario and an extended version incorporating domain shift, which we term \textbf{C}lass \textbf{I}ncremental \textbf{L}earning with \textbf{D}omain shift (\textbf{CILD}).
\begin{itemize}
    \item \textbf{CIL:} Tasks are introduced sequentially without any alteration to the domain, following standard CIL setting.
    \item \textbf{CILD:} Based on CIL, each task $t$ is modified by introducing a unique variation in the domain while sharing the same label space $\mathcal{C}_t$ with CIL. As shown in Fig. ~\ref{fig: Illustration of CILD}, we construct CILD in two ways: \textbf{1)} Synthesizing. We use a pre-trained style transfer model to transfer the original image of CIFAR-100 to multiple domains and use the synthesized dataset DomainCIFAR-100 for training and inference. \textbf{2)}Splitting. We split an existing dataset with domain variation~\cite{domainnet} to form a task sequence. Appendix B.1 shows more details about the CILD scenario setup.
\end{itemize}

\textbf{Evaluation Protocols}. In general, we consider the performance of continual methods from two aspects~\cite{zhangxingxing}: the overall \textbf{A}verage \textbf{A}ccuracy $AA$ of the tasks learned so far, and the \textbf{f}orgetting \textbf{m}easure $FM$ of old tasks. $AA$ evaluates the ability to learn new classes while $FM$ reflects the performance drop of old classes. A lower $FM$ means the model is more robust to fight against forgetting and a higher $AA$ means the model performs well both in learning new knowledge and preserving old knowledge. The detailed mathematical definition of these two metrics can be found in the Appendix A.1.

\begin{table*}[!ht]
    \centering
    
    \setlength{\tabcolsep}{1mm}
    \small{
    \begin{tabular}{cc|cccc|cccc}
    \toprule
                             &                            & \multicolumn{4}{c|}{CIFAR-100}                                                                                                                                            & \multicolumn{4}{c}{DomainNet}                                                                                                                                             \\
    \multirow{-2}{*}{Method} & \multirow{-2}{*}{Scenario} & $AA\uparrow$                           & $FM\downarrow$                                 & $PIV\downarrow$                        & $PFTS\uparrow$                         & $AA\uparrow$                           & $FM\downarrow$                                 & $PIV\downarrow$                        & $PFTS\uparrow$                         \\ \midrule
    iCaRL                    & CIL                        & 58.94                                  & 59.42                                          & 73.50                                  & 23.87                                  & 57.41                                  & 41.27                                          & 74.00                                  & 27.85                                  \\
    iCaRL                    & CILD                       & \cellcolor[HTML]{C0C0C0}\textbf{61.19} & \cellcolor[HTML]{C0C0C0}\textbf{25.98(-33.44)} & \cellcolor[HTML]{C0C0C0}\textbf{56.00} & \cellcolor[HTML]{C0C0C0}\textbf{29.54} & \cellcolor[HTML]{C0C0C0}\textbf{63.32} & \cellcolor[HTML]{C0C0C0}\textbf{22.88(-18.39)} & \cellcolor[HTML]{C0C0C0}\textbf{57.00} & \cellcolor[HTML]{C0C0C0}\textbf{36.11} \\ \midrule
    BiC                      & CIL                        & 70.53                                  & 42.80                                          & 69.00                                  & 56.21                                  & \textbf{62.54}                         & 36.20                                          & 76.50                                  & 54.39                                  \\
    BiC                      & CILD                       & \cellcolor[HTML]{C0C0C0}\textbf{72.16} & \cellcolor[HTML]{C0C0C0}\textbf{16.04(-26.76)} & \cellcolor[HTML]{C0C0C0}\textbf{53.00} & \cellcolor[HTML]{C0C0C0}\textbf{77.25} & \cellcolor[HTML]{C0C0C0}61.82          & \cellcolor[HTML]{C0C0C0}\textbf{13.12(-23.08)} & \cellcolor[HTML]{C0C0C0}\textbf{64.00} & \cellcolor[HTML]{C0C0C0}\textbf{56.03} \\ \midrule
    MEMO                     & CIL                        & \textbf{89.87}                         & 16.91                                          & 72.50                                  & 68.24                                  & \textbf{91.40}                         & 10.67                                          & 73.00                                  & 68.84                                  \\
    MEMO                     & CILD                       & \cellcolor[HTML]{C0C0C0}86.38          & \cellcolor[HTML]{C0C0C0}\textbf{7.56(-9.35)}   & \cellcolor[HTML]{C0C0C0}\textbf{62.50} & \cellcolor[HTML]{C0C0C0}\textbf{95.30} & \cellcolor[HTML]{C0C0C0}86.67          & \cellcolor[HTML]{C0C0C0}\textbf{8.31(-2.36)}   & \cellcolor[HTML]{C0C0C0}\textbf{72.00} & \cellcolor[HTML]{C0C0C0}\textbf{95.30} \\ \midrule
    LwF                      & CIL                        & \textbf{50.49}                         & 49.35                                          & 32.00                                  & 35.80                                  & \textbf{48.62}                         & 46.58                                          & 36.50                                  & 39.65                                  \\
    LwF                      & CILD                       & \cellcolor[HTML]{C0C0C0}48.02          & \cellcolor[HTML]{C0C0C0}\textbf{23.48(-25.87)} & \cellcolor[HTML]{C0C0C0}\textbf{28.50} & \cellcolor[HTML]{C0C0C0}\textbf{46.59} & \cellcolor[HTML]{C0C0C0}45.00          & \cellcolor[HTML]{C0C0C0}\textbf{27.30(-19.28)} & \cellcolor[HTML]{C0C0C0}\textbf{26.00} & \cellcolor[HTML]{C0C0C0}\textbf{43.08} \\ \midrule
    DER                      & CIL                        & 62.51                                  & 40.26                                          & 1.00                                    & 5.11                                   & 67.55                                  & 34.21                                          & 1.00                                    & 6.19                                   \\
    DER                      & CILD                       & \cellcolor[HTML]{C0C0C0}\textbf{72.02} & \cellcolor[HTML]{C0C0C0}\textbf{0.50(-39.76)}  & \cellcolor[HTML]{C0C0C0}1.00            & \cellcolor[HTML]{C0C0C0}\textbf{6.33}  & \cellcolor[HTML]{C0C0C0}\textbf{69.73} & \cellcolor[HTML]{C0C0C0}\textbf{9.56(-24.65)}  & \cellcolor[HTML]{C0C0C0}1.00            & \cellcolor[HTML]{C0C0C0}\textbf{10.57} \\ \midrule
    L2P                      & CIL                        & \textbf{91.36}                         & \textbf{4.42}                                  & 100.00                                 & 30.16                                  & \textbf{91.35}                         & \textbf{3.81}                                  & 100.00                                 & 28.55                                  \\
    L2P                      & CILD                       & \cellcolor[HTML]{C0C0C0}61.37          & \cellcolor[HTML]{C0C0C0}13.46(+9.04)           & \cellcolor[HTML]{C0C0C0}100.00         & \cellcolor[HTML]{C0C0C0}\textbf{59.75} & \cellcolor[HTML]{C0C0C0}83.51          & \cellcolor[HTML]{C0C0C0}5.04(+1.23)            & \cellcolor[HTML]{C0C0C0}100.00         & \cellcolor[HTML]{C0C0C0}\textbf{52.70} \\ \bottomrule
    \end{tabular}
    }
    \caption{Comparative results on average accuracy ($AA$), forgetting ($FM$), interference ($PIV$) and knowledge transfer ($PFTS$) of CIL methods with(in gray rows) and without(in white rows) domain shift on CIFAR-100 and DomainNet. $PIV$ and $PFTS$ will be introduced in section ~\ref{sec: empirical study: quantitative analysis}.}
    \label{tab: CIL vs CILD}
    
\end{table*}

\subsubsection{Observation.}
\label{sec: Empirical Study: observation}
For the baseline CIL methods, we select six representative methods: iCaRL~\cite{icarl}, BiC~\cite{BiC}, MEMO~\cite{memo}, LwF~\cite{lwf}, DER~\cite{DER}, and L2P~\cite{L2P}, covering four different method categories as discussed in sec ~\ref{sec: related works: cil}. Details of these methods' implementation can be found in Appendix C.2.

As shown in Tab. ~\ref{tab: CIL vs CILD}, most CIL methods under CILD demonstrate significantly lower forgetting compared to CIL. This phenomenon is not restricted to a single model or method, and we observe consistent results across various methodologies, regardless of their backbone types or whether pretrained. Note that $AA$ of CILD is generally lower than that of CIL because we use the synthesized images or images belonging to weird domains which may be hard to classify for task $t\geq 2$ in CILD. This leads to a lower initial accuracy of these tasks and a lower $AA$ consequently. If we take a close look at just the first task performance during the whole process, we can observe that the model suffers from much less forgetting under CILD. Details of these observations above can be found in Appendix B.3.

Despite the fantastic low $FM$ of most methods under the CILD scenario, prompt-based method like L2P seems to be the exception, which is more prone to forgetting. This can be attributed to the fact that the backbone remains frozen during the training of prompt-based methods and the key-prompt pairs are the only tunable parameters. These parameters have much less scalability compared to the pre-trained encoder. When faced with examples within the data distribution where the encoder is trained, prompts can adapt to them easily by introducing minor modifications with minor forgetting. But when the input sample style is significantly different from previous ones, these prompts tend to overfit these outlier samples and fail to preserve old knowledge.

\begin{figure}[t]
    \centering
    \includegraphics[width=1.0\linewidth]{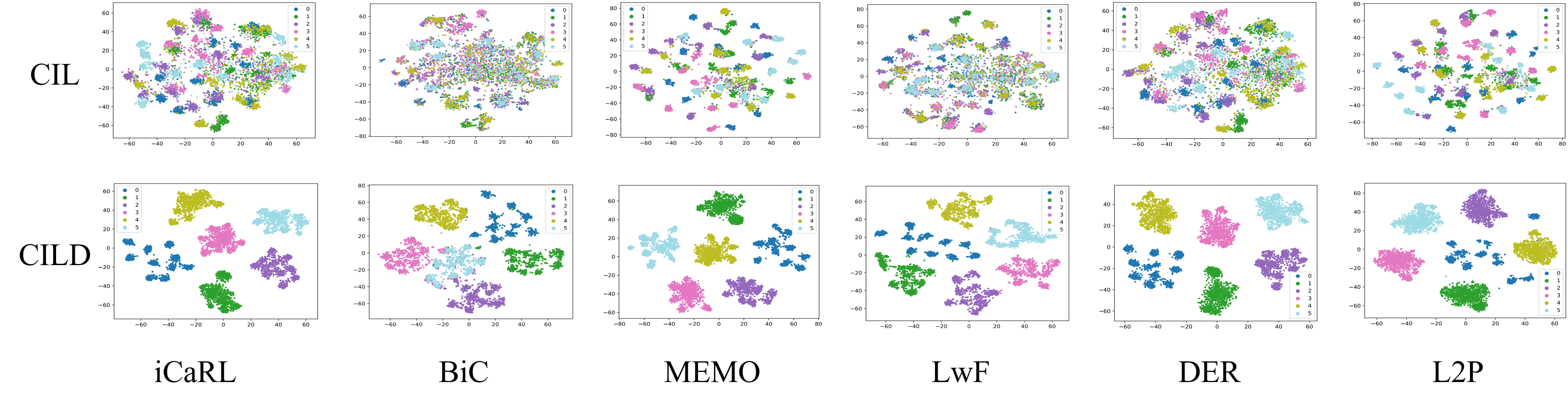}
    \caption{t-SNE visualization of features on CIFAR-100. The top row denotes features extracted by different continual methods under CIL scenario and the bottom row denotes features under CILD scenario. Data points from the same task are marked using the same color.}
    \label{fig: empirical-tsne}
\end{figure}

Moreover, we use t-SNE~\cite{tsne} to visualize the feature representations of each task on CIFAR-100 in Fig. ~\ref{fig: empirical-tsne}. In CIL, there is notable overlap among features of different tasks, indicating a struggle to maintain distinct task-specific features and leading to forgetting. In contrast, in CILD, features from different tasks are separated into well-defined clusters, demonstrating effective preservation of task uniqueness and reduced interference. This suggests that domain shifts help protect the knowledge of previous classes and reduce interference when learning new ones.

\subsection{Quantitative Analysis}
\label{sec: empirical study: quantitative analysis}
Inspired by the surprisingly low forgetting rate of CILD, one hypothesis is that just certain parts of the parameters are updated conditioned on the distinguishable input representations, thus reducing the interference between tasks. To quantify the interference among parameters that are heavily updated across different tasks, we design a metric named $PIV$ (\textbf{P}arameter \textbf{I}nterference \textbf{V}alue) to capture how changes in one set of parameters affect the others across the sequence of tasks, and a metric named $PFTS$(\textbf{P}arameter \textbf{F}orward \textbf{T}ransfer \textbf{S}core) to capture how much knowledge is relatively learned by model. For clarity, the mathematical definition of these two metrics is presented in Appendix A.2.

\begin{figure*}[!t]
    \begin{center}
    \includegraphics[width=2.0\columnwidth]{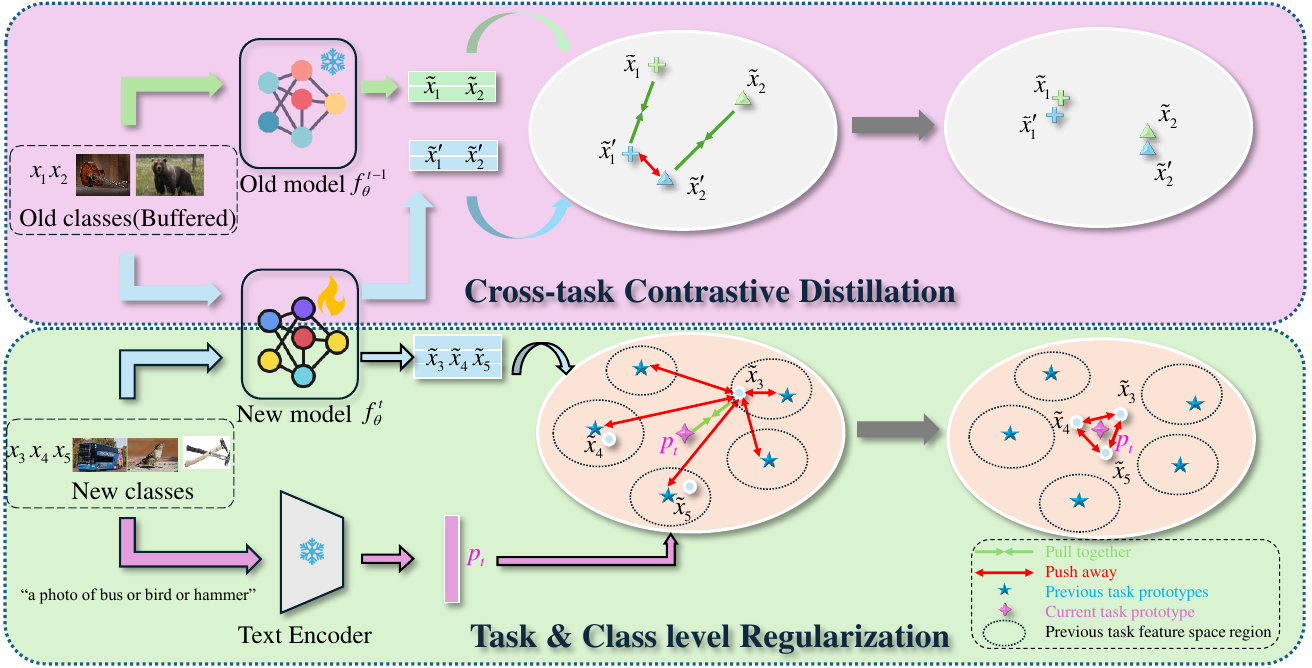}
    \end{center}
       \caption{
            The overview framework of \methodname. \methodname includes Task\&Class -level Regularization and Cross-task Contrastive Distillation. In Task\&Class -level Regularization, samples in the current task are pulled toward the current task prototype while pushed away from previous task prototypes, leading to a discriminative feature distribution away from other tasks. Cross-task Contrastive Distillation helps align current model with previous one and preserve the features of old classes.
        }
    \label{fig: method_overview}
\end{figure*}

The last two columns in Tab. ~\ref{tab: CIL vs CILD} demonstrate the $PIV$ and $PFTS$ of various continual methods under CIL and CILD. It is shown that \textbf{1) }Methods under CILD demonstrate lower $PIV$ compared to CIL, suggesting that domain shift helps reduce the interference across tasks. \textbf{2)} Methods under CILD usually demonstrate higher $PFTS$, indicating that domain shift may be beneficial for the model to learn various distinguishable representations in an ordered way. For architecture-based method DER, $PIV$ under two settings are relatively lower because DER introduces new layers responsible for learning each task respectively. For prompt-based method L2P, $PIV$ is 100\% because a limited number of prompts are the only parameters that can be updated during training. At the same time, the $PFTS$ is much higher than that under CIL, demonstrating the strong learning ability of the shared prompts. This leads to high interference across tasks in L2P and makes it perform worse under CILD.



\section{Method}

In section ~\ref{sec: empirical_study}, we observed that domain shifts across tasks could significantly enhance the method's resistance to forgetting in class-incremental learning. Integrating these domain shifts into existing CIL methods to boost performance is a natural idea. However, this raises a challenge: the model cannot determine which domain shift to apply to an input during inference. This resembles the paradigm of task-incremental learning~\cite{zhangxingxing}, where the task ID (indicating the domain shift in our case) is needed at inference, which contradicts the CIL setting.

Instead, inspired by our observation, we propose to promote the differentiation of task-specific features and aim to simulate the beneficial effects of domain shift within the feature space rather than the input. Consequently, we introduce \textbf{\methodname} (\textbf{Dis}tinguishable feature for \textbf{Co}ntinual Learning), a simple yet effective rehearsal-based method employing contrastive learning to learn distinguishable representations for continual tasks. \methodname is built upon existing rehearsal-based baseline continual methods and can be easily incorporated into them. The core idea of \methodname is to enforce a margin between the feature distributions of the current categories and prototypes of previous categories, thereby preventing the interference of old knowledge. As shown in Fig. ~\ref{fig: method_overview}, \methodname comprises mainly two parts: task-level and class-level regularization and cross-task contrastive distillation.

\textbf{Prototype construction:} \label{sec: prototype construction}In our work, we try two types of prototype: text prototype and image feature prototype. For a class-incremental task at stage \(t\), given a mini-batch with index $i$ of $N$ training samples \({D}_t^i=\{(x_j,\ y_j)\}_{j=1}^N\), where \(x_j\in \mathbb{R}^{3\times H\times W}\) and \(y_j\in \mathbb{R}\) denote the image and numerical class label of the \(j\)-th sample respectively. We feed $\{x_j\}_{j=1}^N$ into feature extractor $\mathbf{F}_\theta ^t$ to get their features $\{\hat{x_j}\}_{j=1}^N\in\mathbb{R}^{N\times D}$, followed by a projector $\psi_t$ to project into a lower dimension space $\{\Tilde{x_j}\}_{j=1}^N\in\mathbb{R}^{N\times d}$. The batch-wise image feature prototype $p_i$ is then calculated as $\frac 1N\sum_{j=1}^N\Tilde{x_j}$. 

For batch-wise text prototype, we collect all class names of this batch and feed the below text prompt to CLIP~\cite{clip} text encoder to extract embedding $p_i$:
\begin{center}
    ``\textit{a photo of \{class 0\} or \{class 1\} or ... or \{class n\}}"
\end{center}

To let the batch-wise $p_i$ approximate prototype $P_i$ of the whole task, we calculate the momentum accumulation of $p_i$: 
{
\begin{align}
    p_i= \frac {i-1}{i}p_{i-1}+\frac {1}{i}p_i
\end{align}
}
After all $N_t$ mini-batches of task $t$, $p_{N_t}$ is stored in the prototype pool $\mathbf{P}$ as the prototype $P_t$ of task $t$.

\subsection{Task-level and Class-level Regularization}
In task-level regularization, we aim to keep features of the current task away from the prototypes of previous tasks. 

We treat each sample $x_j$ in the $i$-th mini-batch as the anchor and treat the prototype $p_i$ of this batch as the only positive sample \(x_{p_i}\). Prototypes of all previous tasks $P_k,\ \forall k< t$ are negative samples. As a result, $N\times (t-1)$ pairs of triplets can be constructed in this mini-batch, and then task-level contrastive loss $\mathcal{L}_{tcon}$ is:
{\small
\begin{align}
    \label{eq: tcon}
    \mathcal{L}_{tcon} =\frac{1}{N\times (t-1)}\sum_{j=1}^{N}\sum_{k=1}^{t-1}Triplet(x_j, p_i, P_k)\\
    Triplet(a, p, n) = \log \left(1+\exp( 1-S(a,p)+S(a,n)\right))
\end{align}
}
where $S(x,y)$ is the cosine similarity function. $\mathcal{L}_{tcon}$ ensures the feature space margin between the current task and previous ones, preventing task interference.

In class-level regularization, the goal is to learn discriminative features for each class in the current task. For each sample $x_j$, we randomly select a sample having the same label with $x_j$ in this batch as the positive sample $x_p$, and select a random negative sample $x_n$ having a different label. Then the class-level contrastive loss $\mathcal{L}_{ccon}$ is:
{
\begin{align}
    \mathcal{L}_{ccon} = \frac 1N\sum_{j=1}^N Triplet(x_j, x_p, x_n)
\end{align}
}

\subsection{Cross-task Contrastive Distillation}
Despite the satisfactory performance achieved through the combined use of the aforementioned regularization and the rehearsal mechanism in the baseline method, the incorporation of an explicit mechanism to preserve acquired knowledge could yield further benefits. Here we leverage a \textbf{C}ross-task \textbf{C}ontrastive \textbf{D}istillation loss(\textbf{CCD}) to help align the current student model with the old teacher model. At task $t$, we copy the trained model of task $t-1$ as the teacher model $f_{\theta}^{t-1}$ which remains frozen during task $t$, and the current model $f_\theta^t$ serves as the student model. The contrastive distillation aims to align the feature representation of the student model and teacher model for old classes, preventing the degradation of old classes' features. Specifically, for a sample $x_j$ in the rehearsal sample set $\mathcal{R}=\{(x_j, y_j)|y_j\notin\mathcal{C}_t\}$, we use the teacher model $f_\theta^{t-1}$ and student model $f_\theta^t$ to extract their features, denoted as $\tilde{x_j}$ and $\tilde{x_j}'$. The contrastive distillation loss is written as:
{
\begin{align}
    \mathcal{L}_{ccd} = \sum_{(x_j,y_j)\in\mathcal{R}}\sum_{k\in\mathcal{R},y_k\neq y_j}Triplet(\tilde{x_j}', \tilde{x_j}, \tilde{x_k}')
\end{align}
}
$\mathcal{L}_{ccd}$ places restrictions on current student model to distill knowledge of the same old class from teacher model and simultaneously keep away from other different old classes.

\subsection{Generalize to Various Types of CIL Methods}
\methodname not only works well on rehearsal-based continual methods but also can generalize to other categories, such as regularization-based and prompt-based methods. For regularization-based methods, \methodname can be directly incorporated discarding CCD with minor performance degradation. For prompt-based methods such as L2P~\cite{L2P}, we just need to impose regularization on selected prompt keys like Eq. ~\ref{eq: tcon} instead of image features. More details about incorporating \methodname into prompt-based methods can be found in Appendix C.3.

As a result, the total loss $\mathcal{L}$ of \methodname is:
\begin{align}
    \mathcal{L}_{baseline} + \lambda_{tcon}\mathcal{L}_{tcon} + \lambda_{ccon}\mathcal{L}_{ccon} + \lambda_{ccd}\mathcal{L}_{ccd}
    \label{eq: loss}
\end{align}
where $\mathcal{L}_{baseline}$ refers to the vanilla loss of the baseline continual method, $\lambda_{tcon}$, $\lambda_{ccon}$ and $\lambda_{ccd}$ are hyperparameter to balance these losses. In our experiments, $\lambda_{tcon}$ and $\lambda_{ccon}$ are set to 0.5 and $\lambda_{ccd}=1$. Evaluations of other possible values are reported in Appendix C.5.

\section{Experiments}

\subsection{Experiment Setup}
\label{sec: exp: experiment setup}

\textbf{Datasets and Implementation:} \label{sec: experiment: implementation}We perform experiments on CIFAR100, Fashion-MNIST, and Tiny-ImageNet. Details of these datasets and continual task split are in Appendix C.1.

Our code is implemented in PyTorch and based on LAMDA-PILOT\cite{pilot1}, which is an open-source framework for easily designing continual methods. We select iCaRL\cite{icarl}, BiC~\cite{BiC}, LwF~\cite{lwf}, DER\cite{DER}, and L2P\cite{L2P} as baseline methods and integrate DisCo into them. We also compare our methods with the Co2L~\cite{co2l}, which is a rehearsal-based method leveraging contrastive loss to learn stable representations. Appendix C.2 shows more details about these methods' implementations. For both ResNet~\cite{resnet} and ViT~\cite{vit} -based models, we train all tasks for 100 epochs, 60-th and 80-th epochs being milestones. For models built on ResNet, we set weight decay $w=5e-4$ and learning rate $lr=0.1$ with $\times 0.1$ at milestones. For ViT-based models, $w=2e-4$, $lr=1e-3$ with $\times 0.1$ at milestones.

\begin{table}[t]
    \centering
    
    \setlength{\tabcolsep}{0.5mm}
    \small{
    \begin{tabular}{c|cc|cc|cc}
    \toprule
    \multirow{2}{*}{Method}  & \multicolumn{2}{c}{CIFAR100}   & \multicolumn{2}{c}{FashionMNIST} & \multicolumn{2}{c}{TinyImageNet} \\
                                                              & $AA\uparrow$   & $FM\downarrow$ & $AA\uparrow$    & $FM\downarrow$  & $AA\uparrow$    & $FM\downarrow$  \\ \midrule
    iCaRL                          & 64.24          & 51.34          & 69.41           & 47.44           & 8.57            & 81.40           \\
    iCaRL + DisCo-I                                             & \textbf{70.11} & \textbf{33.96} & \textbf{72.57}  & 34.45           & 10.87           & \textbf{70.19}  \\
    iCaRL + DisCo-T                                             & 63.35          & 35.26          & 70.69           & \textbf{33.65}  & \textbf{11.88}  & 70.46           \\ \cline{2-7}
    BiC                                                      & 67.04          & 46.51          & 73.63           & 38.77           & \textbf{8.42}   & 78.42           \\
    BiC + Disco-I                                               & \textbf{69.89} & 28.54          & 73.24           & \textbf{34.22}  & 8.16            & 68.05           \\
    BiC + Disco-T                                               & 67.68          & \textbf{23.08} & \textbf{74.18}  & 35.00           & 8.33            & \textbf{67.94}  \\ \cline{2-7}
    Co2L${^\dagger}$                                          & 71.25          & 32.17          & 68.54           & 34.66           & 14.02           & 74.55           \\ \midrule
    LwF                        & 51.30          & 55.98          & 53.15           & 51.27           & 5.06            & 85.50           \\
    LwF + DisCo-I                                               & 56.42          & 36.52          & \textbf{57.14}  & \textbf{37.69}  & \textbf{7.87}   & 79.58           \\
    LwF + DisCo-T                                               & \textbf{56.87} & \textbf{33.89} & 52.87           & 44.11           & 6.22            & \textbf{77.15}  \\ \midrule
    DER                         & 63.91          & 40.18          & 68.21           & 37.85           & 11.58           & 77.99           \\
    DER + Disco-I                                               & 64.87          & 36.41          & \textbf{72.41}  & \textbf{29.86}  & 12.08           & \textbf{72.11}  \\
    DER + Disco-T                                               & \textbf{69.51} & \textbf{33.74} & 70.59           & 31.43           & \textbf{12.86}  & 73.02           \\ \midrule
    L2P                               & 82.65          & 7.62           & 85.21           & 6.74            & 29.54           & 44.30           \\
    L2P + Disco-I                                               & 82.78          & 7.98           & \textbf{86.15}  & 7.88            & 28.43           & 43.32           \\
    L2P + Disco-T                                               & \textbf{83.12} & \textbf{6.80}  & 84.99           & \textbf{6.07}   & \textbf{31.88}  & \textbf{40.00} \\
    \bottomrule
    \end{tabular}
    }
    
    \caption{Main result of incorporating \methodname into existing continual methods. We group the compared methods by their category(rehearsal-based, regularization-based, architecture-based, or prompt-based). Co2L${^\dagger}$ is a compared rehearsal-based method leveraging contrastive loss. Disco-I means image features as prototypes and Disco-T means text features as prototypes, as described in section ~\ref{sec: prototype construction}. Each result is averaged over 3 runs.}
    \label{tab: Main Result}
\end{table}

\begin{table*}[!t]
    \centering
    
    \setlength{\tabcolsep}{1mm}
    \small{
    \begin{tabular}{cccc|ccc|ccc|ccc}
        \toprule
        \multirow{2}{*}{Method} & \multicolumn{3}{c|}{\methodname}               & \multicolumn{3}{c|}{\methodname w/o Ccon}    & \multicolumn{3}{c|}{\methodname w/o Tcon}      & \multicolumn{3}{c}{\methodname w/o CCD}      \\
                                & $AA\uparrow$   & $FM\downarrow$ & $IA\uparrow$ & $AA\uparrow$ & $FM\downarrow$ & $IA\uparrow$ & $AA\uparrow$   & $FM\downarrow$ & $IA\uparrow$ & $AA\uparrow$ & $FM\downarrow$ & $IA\uparrow$ \\ \midrule
        iCaRL + DisCo-I           & \textbf{70.11} & 33.96          & 80.16        & 67.21        & \textbf{32.15} & 75.22        & 66.00          & 47.41          & 82.54        & 69.35        & 35.47          & 80.34        \\
        BiC + DisCo-I             & \textbf{69.89} & 28.54          & 75.21        & 63.32        & \textbf{28.16} & 72.52        & 65.15          & 44.25          & 74.67        & 67.33        & 26.68          & 76.10        \\
        LwF + DisCo-I             & \textbf{56.42} & 36.52          & 81.67        & 51.34        & \textbf{35.99} & 76.02        & 52.88          & 52.63          & 81.24        & -            & -              & -            \\
        DER + DisCo-I             & \textbf{64.87} & 36.41          & 78.56        & 60.35        & 37.68          & 74.48        & 61.23          & 42.16          & 79.51        & 62.76        & \textbf{35.16} & 78.68        \\
        L2P + DisCo-T             & 83.12          & \textbf{6.80}  & 85.43        & 81.16        & 6.75           & 84.36        & \textbf{83.16} & 7.87           & 86.49        & -            & -              & -           \\
        \bottomrule
    \end{tabular}
    }
    
    \caption{Ablation study on \methodname components on CIFAR-100. $IA$ refers to the average \textbf{I}nitial \textbf{A}ccuracy of each task as in Appendix A.1. Tcon refers to task-level regularization, Ccon refers to class-level regularization and CCD refers to cross-task distillation. For each method row, we highlight the highest $AA$ and lowest $FM$. For non-rehearsal-based methods, there is no ablation on CCD, which is marked as ``-".}
    \label{tab: ablation study on modules}
\end{table*}
\subsection{Evaluation on Three Benchmarks}
Tab.~\ref{tab: Main Result} shows the result of incorporating \methodname into existing continual methods. It can be observed that \methodname helps most methods alleviate forgetting ($FM$) and improves average accuracy ($AA$). Especially, \methodname can boost the performance of rehearsal-based methods significantly: increase $AA$ by 5.87\% and reduce $FM$ by 17.38\% for iCaRL on CIFAR-100, increase $AA$ by 3.31\% and reduce $FM$ by 10.94\% on Tiny-ImageNet. Moreover, plugging \methodname achieves comparable or even better performance compared to the contrastive-based related work Co2L~\cite{co2l}. For regularization-based methods, \methodname works as well leveraging our task\&class -level and their intrinsic regularization module, reducing $FM$ by 22.09\% for LwF on CIFAR-100 and 8.35\% on Tiny-ImageNet.

Compared to image features as prototypes (\methodname-I), using task text features as prototypes (\methodname-T) performs better on prompt-based methods. There are several possible reasons: \textbf{1)} L2P employs a prompt pool that serves as a bridge between the task's data and the model, guiding the model's focus toward the most relevant features for each task. Text features encapsulate high-level semantic information, and can effectively steer the model's attention to conceptual similarities and distinctions between classes or tasks. \textbf{2)}Prompt-based methods are inherently more flexible in handling text features since they often originate from NLP backgrounds. Thus, they can more effectively utilize text prototypes to guide the learning process across different tasks.

The improvement of average accuracy is smaller than that of the forgetting rate, partly because task-level regularization imposes stricter restrictions on the feature distributions, making it a little more difficult to classify new classes, i.e. low initial accuracy of each task. Ablation study of these regularization modules is shown in section ~\ref{sec: ablation study on modules}. More additional results such as t-sne visualization and other findings are presented in Appendix C.4.

\subsection{Ablation Study}

\subsubsection{Ablation study on components of \methodname}
\label{sec: ablation study on modules}
We conduct an ablation study of task\&class -level regularization and CCD on CIFAR-100 as shown in Tab. ~\ref{tab: ablation study on modules}.  Task level regularization (Tcon) plays the most important role in helping the model distinguish different tasks, greatly reducing the interference and forgetting rate $FM$. However, only using Tcon makes it difficult for the model to learn new knowledge, leading to a lower $IA$ and a lower $AA$ consequently. Class level regularization (Ccon) helps the model learn discriminative features for different classes in the same task, thus making it easier to learn new tasks i.e., higher $IA$, contributing to a higher $AA$ together with Tcon. Moreover, we find that CCD also greatly helps maintain old class features and prototype distributions, leading to lower $FM$ and higher $AA$.

\subsubsection{Ablation study on task length}
\label{sec: ablation study on task length}
\begin{figure}[t]
  \centering
  \includegraphics[width=1.0\columnwidth]{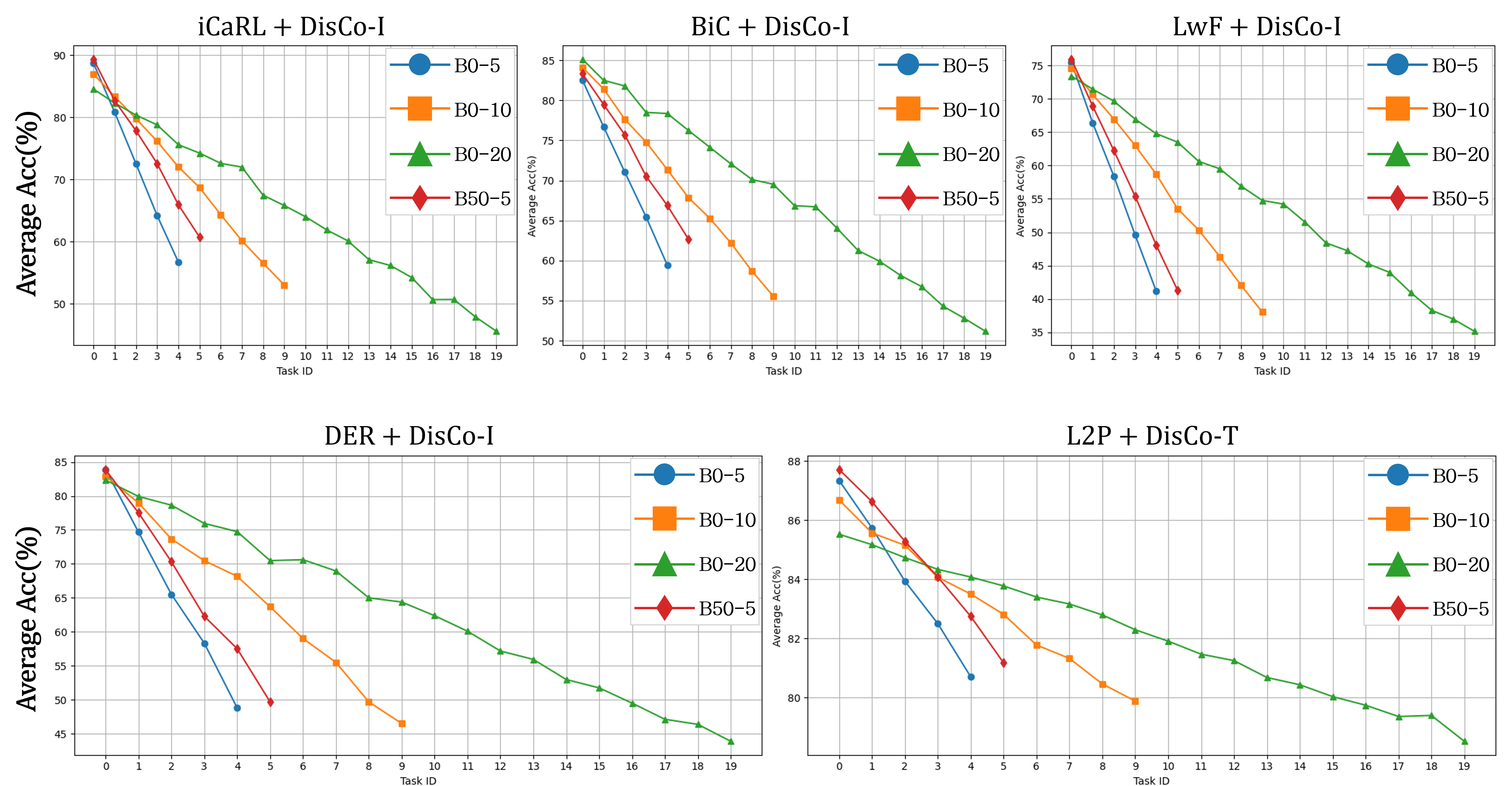}
    \caption{Ablation study on incremental task length. $B\{X\}$-$\{Y\}$ means there are $X$ classes in task 0 and the rest are evenly distributed in $Y$ tasks. The y-axis means the $AA_k$ at task $k$.}
    \label{fig: ablation on task length}
\end{figure}

We conduct experiments on different increment strategies on CIFAR-100. We separate these 100 classes into several groups as most works do\cite{icarl, L2P}. Fig. ~\ref{fig: ablation on task length} demonstrates that the forgetting trend is relatively milder as task length grows. This means \methodname can achieve good performance under different task length situations. Moreover, the forgetting trend of B50-5 is smaller than that of others, because \methodname relies on prototypes to guide the distribution of new tasks, the greater the sample number of each task is, the better performance \methodname achieves. 

\section{Conclusion}
This paper demonstrates a counter-intuitive phenomenon: incorporating domain shift into class-incremental tasks significantly reduces catastrophic forgetting. Inspired by this, we propose \methodname, a contrastive learning-based simple yet effective plug-and-play method that effectively promotes distinct feature distributions for each task, mitigating interference and enhancing performance. Experimental results on various datasets show that \methodname can be easily integrated into other continual methods to boost performance.

\section*{Acknowledgements}
This work is supported in part by the National Natural Science Foundation of China under Grant 62476054, and in part by the Fundamental Research Funds for the Central Universities of China. This research work is supported by the Big Data Computing Center of Southeast University.

\bibliography{ref}

\begin{thebibliography}{45}
\providecommand{\natexlab}[1]{#1}

\bibitem[{Buzzega et~al.(2020)Buzzega, Boschini, Porrello, Abati, and Calderara}]{DER}
Buzzega, P.; Boschini, M.; Porrello, A.; Abati, D.; and Calderara, S. 2020.
\newblock Dark experience for general continual learning: a strong, simple baseline.
\newblock \emph{Advances in neural information processing systems}, 33: 15920--15930.

\bibitem[{Caccia et~al.(2020)Caccia, Belilovsky, Caccia, and Pineau}]{AQM}
Caccia, L.; Belilovsky, E.; Caccia, M.; and Pineau, J. 2020.
\newblock Online learned continual compression with adaptive quantization modules.
\newblock In \emph{International conference on machine learning}, 1240--1250. PMLR.

\bibitem[{Cha, Lee, and Shin(2021)}]{co2l}
Cha, H.; Lee, J.; and Shin, J. 2021.
\newblock Co2l: Contrastive continual learning.
\newblock In \emph{Proceedings of the IEEE/CVF International conference on computer vision}, 9516--9525.

\bibitem[{Deng et~al.(2009)Deng, Dong, Socher, Li, Li, and Fei-Fei}]{imagenet}
Deng, J.; Dong, W.; Socher, R.; Li, L.-J.; Li, K.; and Fei-Fei, L. 2009.
\newblock Imagenet: A large-scale hierarchical image database.
\newblock In \emph{2009 IEEE conference on computer vision and pattern recognition}, 248--255. Ieee.

\bibitem[{Dosovitskiy et~al.(2020)Dosovitskiy, Beyer, Kolesnikov, Weissenborn, Zhai, Unterthiner, Dehghani, Minderer, Heigold, Gelly et~al.}]{vit}
Dosovitskiy, A.; Beyer, L.; Kolesnikov, A.; Weissenborn, D.; Zhai, X.; Unterthiner, T.; Dehghani, M.; Minderer, M.; Heigold, G.; Gelly, S.; et~al. 2020.
\newblock An image is worth 16x16 words: Transformers for image recognition at scale.
\newblock \emph{arXiv preprint arXiv:2010.11929}.

\bibitem[{Golkar, Kagan, and Cho(2019)}]{architecture1}
Golkar, S.; Kagan, M.; and Cho, K. 2019.
\newblock Continual learning via neural pruning.
\newblock \emph{arXiv preprint arXiv:1903.04476}.

\bibitem[{He et~al.(2016)He, Zhang, Ren, and Sun}]{resnet}
He, K.; Zhang, X.; Ren, S.; and Sun, J. 2016.
\newblock Deep residual learning for image recognition.
\newblock In \emph{Proceedings of the IEEE conference on computer vision and pattern recognition}, 770--778.

\bibitem[{Jaccard(1901)}]{jaccard1901etude}
Jaccard, P. 1901.
\newblock {\'E}tude comparative de la distribution florale dans une portion des Alpes et des Jura.
\newblock \emph{Bull Soc Vaudoise Sci Nat}, 37: 547--579.

\bibitem[{Kirkpatrick et~al.(2017)Kirkpatrick, Pascanu, Rabinowitz, Veness, Desjardins, Rusu, Milan, Quan, Ramalho, Grabska-Barwinska et~al.}]{ewc}
Kirkpatrick, J.; Pascanu, R.; Rabinowitz, N.; Veness, J.; Desjardins, G.; Rusu, A.~A.; Milan, K.; Quan, J.; Ramalho, T.; Grabska-Barwinska, A.; et~al. 2017.
\newblock Overcoming catastrophic forgetting in neural networks.
\newblock \emph{Proceedings of the national academy of sciences}, 114(13): 3521--3526.

\bibitem[{Krizhevsky and Hinton(2009)}]{cifar100}
Krizhevsky, A.; and Hinton, G. 2009.
\newblock Learning multiple layers of features from tiny images.
\newblock \emph{Handbook of Systemic Autoimmune Diseases}, 1(4).

\bibitem[{Kundu et~al.(2020)Kundu, Venkatesh, Venkat, Revanur, and Babu}]{UCL1}
Kundu, J.~N.; Venkatesh, R.~M.; Venkat, N.; Revanur, A.; and Babu, R.~V. 2020.
\newblock Class-incremental domain adaptation.
\newblock In \emph{Computer Vision--ECCV 2020: 16th European Conference, Glasgow, UK, August 23--28, 2020, Proceedings, Part XIII 16}, 53--69. Springer.

\bibitem[{Lai et~al.(2024)Lai, Zhou, Liu, and Zhou}]{lai_domainshift}
Lai, Y.; Zhou, Y.; Liu, X.; and Zhou, T. 2024.
\newblock Memory-Assisted Sub-Prototype Mining for Universal Domain Adaptation.
\newblock In \emph{The Twelfth International Conference on Learning Representations}.

\bibitem[{Li and Hoiem(2017)}]{lwf}
Li, Z.; and Hoiem, D. 2017.
\newblock Learning without forgetting.
\newblock \emph{IEEE transactions on pattern analysis and machine intelligence}, 40(12): 2935--2947.

\bibitem[{Lin, Chu, and Lai(2022)}]{regularization-weight}
Lin, G.; Chu, H.; and Lai, H. 2022.
\newblock Towards better plasticity-stability trade-off in incremental learning: A simple linear connector.
\newblock In \emph{Proceedings of the IEEE/CVF Conference on Computer Vision and Pattern Recognition}, 89--98.

\bibitem[{Liu et~al.(2020)Liu, Wu, Menta, Herranz, Raducanu, Bagdanov, Jui, and de~Weijer}]{generative1}
Liu, X.; Wu, C.; Menta, M.; Herranz, L.; Raducanu, B.; Bagdanov, A.~D.; Jui, S.; and de~Weijer, J.~v. 2020.
\newblock Generative feature replay for class-incremental learning.
\newblock In \emph{Proceedings of the IEEE/CVF Conference on Computer Vision and Pattern Recognition Workshops}, 226--227.

\bibitem[{Liu and Zhou(2024)}]{liu_domainshift}
Liu, X.; and Zhou, Y. 2024.
\newblock COCA: Classifier-Oriented Calibration via Textual Prototype for Source-Free Universal Domain Adaptation.
\newblock In \emph{Proceedings of the Asian Conference on Computer Vision}, 1671--1687.

\bibitem[{Lopez-Paz and Ranzato(2017)}]{AA}
Lopez-Paz, D.; and Ranzato, M. 2017.
\newblock Gradient episodic memory for continual learning.
\newblock \emph{Advances in neural information processing systems}, 30.

\bibitem[{Luo et~al.(2024)Luo, Chen, Tian, Liu, Hou, Zhang, Shen, Wu, Geng, Zhou et~al.}]{survey_ad}
Luo, S.; Chen, W.; Tian, W.; Liu, R.; Hou, L.; Zhang, X.; Shen, H.; Wu, R.; Geng, S.; Zhou, Y.; et~al. 2024.
\newblock Delving into Multi-modal Multi-task Foundation Models for Road Scene Understanding: From Learning Paradigm Perspectives.
\newblock \emph{IEEE Transactions on Intelligent Vehicles}.

\bibitem[{Mallya and Lazebnik(2018)}]{architecture2}
Mallya, A.; and Lazebnik, S. 2018.
\newblock Packnet: Adding multiple tasks to a single network by iterative pruning.
\newblock In \emph{Proceedings of the IEEE conference on Computer Vision and Pattern Recognition}, 7765--7773.

\bibitem[{McCloskey and Cohen(1989)}]{catastrophic1}
McCloskey, M.; and Cohen, N.~J. 1989.
\newblock Catastrophic interference in connectionist networks: The sequential learning problem.
\newblock In \emph{Psychology of learning and motivation}, volume~24, 109--165. Elsevier.

\bibitem[{Peng et~al.(2019)Peng, Bai, Xia, Huang, Saenko, and Wang}]{domainnet}
Peng, X.; Bai, Q.; Xia, X.; Huang, Z.; Saenko, K.; and Wang, B. 2019.
\newblock Moment matching for multi-source domain adaptation.
\newblock In \emph{Proceedings of the IEEE/CVF international conference on computer vision}, 1406--1415.

\bibitem[{Pham, Liu, and Hoi(2021)}]{dualnet}
Pham, Q.; Liu, C.; and Hoi, S. 2021.
\newblock Dualnet: Continual learning, fast and slow.
\newblock \emph{Advances in Neural Information Processing Systems}, 34: 16131--16144.

\bibitem[{Radford et~al.(2021)Radford, Kim, Hallacy, Ramesh, Goh, Agarwal, Sastry, Askell, Mishkin, Clark et~al.}]{clip}
Radford, A.; Kim, J.~W.; Hallacy, C.; Ramesh, A.; Goh, G.; Agarwal, S.; Sastry, G.; Askell, A.; Mishkin, P.; Clark, J.; et~al. 2021.
\newblock Learning transferable visual models from natural language supervision.
\newblock In \emph{International conference on machine learning}, 8748--8763. PMLR.

\bibitem[{Razdaibiedina et~al.(2023)Razdaibiedina, Mao, Hou, Khabsa, Lewis, and Almahairi}]{progressiveprompt}
Razdaibiedina, A.; Mao, Y.; Hou, R.; Khabsa, M.; Lewis, M.; and Almahairi, A. 2023.
\newblock Progressive prompts: Continual learning for language models.
\newblock \emph{arXiv preprint arXiv:2301.12314}.

\bibitem[{Rebuffi, Kolesnikov, and Lampert(2016)}]{icarl}
Rebuffi, S.; Kolesnikov, A.; and Lampert, C.~H. 2016.
\newblock icarl: Incremental classifier and representation learning. CoRR abs/1611.07725 (2016).
\newblock \emph{arXiv preprint arXiv:1611.07725}.

\bibitem[{Rusu et~al.(2016)Rusu, Rabinowitz, Desjardins, Soyer, Kirkpatrick, Kavukcuoglu, Pascanu, and Hadsell}]{architecture3}
Rusu, A.~A.; Rabinowitz, N.~C.; Desjardins, G.; Soyer, H.; Kirkpatrick, J.; Kavukcuoglu, K.; Pascanu, R.; and Hadsell, R. 2016.
\newblock Progressive neural networks.
\newblock \emph{arXiv preprint arXiv:1606.04671}.

\bibitem[{Sheng et~al.(2018)Sheng, Lin, Shao, and Wang}]{avatarnet}
Sheng, L.; Lin, Z.; Shao, J.; and Wang, X. 2018.
\newblock Avatar-net: Multi-scale zero-shot style transfer by feature decoration.
\newblock In \emph{Proceedings of the IEEE conference on computer vision and pattern recognition}, 8242--8250.

\bibitem[{Simon et~al.(2022)Simon, Faraki, Tsai, Yu, Schulter, Suh, Harandi, and Chandraker}]{UCL2}
Simon, C.; Faraki, M.; Tsai, Y.-H.; Yu, X.; Schulter, S.; Suh, Y.; Harandi, M.; and Chandraker, M. 2022.
\newblock On generalizing beyond domains in cross-domain continual learning.
\newblock In \emph{Proceedings of the IEEE/CVF Conference on Computer Vision and Pattern Recognition}, 9265--9274.

\bibitem[{Smith et~al.(2023)Smith, Karlinsky, Gutta, Cascante-Bonilla, Kim, Arbelle, Panda, Feris, and Kira}]{codaprompt}
Smith, J.~S.; Karlinsky, L.; Gutta, V.; Cascante-Bonilla, P.; Kim, D.; Arbelle, A.; Panda, R.; Feris, R.; and Kira, Z. 2023.
\newblock Coda-prompt: Continual decomposed attention-based prompting for rehearsal-free continual learning.
\newblock In \emph{Proceedings of the IEEE/CVF Conference on Computer Vision and Pattern Recognition}, 11909--11919.

\bibitem[{Sun et~al.(2023)Sun, Zhou, Ye, and Zhan}]{pilot1}
Sun, H.-L.; Zhou, D.-W.; Ye, H.-J.; and Zhan, D.-C. 2023.
\newblock PILOT: A Pre-Trained Model-Based Continual Learning Toolbox.
\newblock \emph{arXiv preprint arXiv:2309.07117}.

\bibitem[{Tang et~al.(2021)Tang, Su, Chen, and Ouyang}]{DIL1}
Tang, S.; Su, P.; Chen, D.; and Ouyang, W. 2021.
\newblock Gradient regularized contrastive learning for continual domain adaptation.
\newblock In \emph{Proceedings of the AAAI Conference on Artificial Intelligence}, volume~35, 2665--2673.

\bibitem[{Tao et~al.(2020)Tao, Hong, Chang, and Gong}]{DIL2}
Tao, X.; Hong, X.; Chang, X.; and Gong, Y. 2020.
\newblock Bi-Objective Continual Learning: Learning ‘New’ While Consolidating ‘Known’.
\newblock \emph{Proceedings of the AAAI Conference on Artificial Intelligence}, 34(04): 5989--5996.

\bibitem[{Van~de Ven, Siegelmann, and Tolias(2020)}]{generative2}
Van~de Ven, G.~M.; Siegelmann, H.~T.; and Tolias, A.~S. 2020.
\newblock Brain-inspired replay for continual learning with artificial neural networks.
\newblock \emph{Nature communications}, 11(1): 4069.

\bibitem[{Van~der Maaten and Hinton(2008)}]{tsne}
Van~der Maaten, L.; and Hinton, G. 2008.
\newblock Visualizing data using t-SNE.
\newblock \emph{Journal of machine learning research}, 9(11).

\bibitem[{Volpi, Larlus, and Rogez(2021)}]{DIL3}
Volpi, R.; Larlus, D.; and Rogez, G. 2021.
\newblock Continual adaptation of visual representations via domain randomization and meta-learning.
\newblock In \emph{Proceedings of the IEEE/CVF Conference on Computer Vision and Pattern Recognition}, 4443--4453.

\bibitem[{Wah et~al.(2011)Wah, Branson, Welinder, Perona, and Belongie}]{cub}
Wah, C.; Branson, S.; Welinder, P.; Perona, P.; and Belongie, S. 2011.
\newblock The caltech-ucsd birds-200-2011 dataset.

\bibitem[{Wang et~al.(2024)Wang, Zhang, Su, and Zhu}]{zhangxingxing}
Wang, L.; Zhang, X.; Su, H.; and Zhu, J. 2024.
\newblock A comprehensive survey of continual learning: Theory, method and application.
\newblock \emph{IEEE Transactions on Pattern Analysis and Machine Intelligence}.

\bibitem[{Wang et~al.(2022{\natexlab{a}})Wang, Liu, Duan, and Tao}]{regularization-function}
Wang, Z.; Liu, L.; Duan, Y.; and Tao, D. 2022{\natexlab{a}}.
\newblock Continual learning through retrieval and imagination.
\newblock In \emph{Proceedings of the AAAI Conference on Artificial Intelligence}, 8, 8594--8602.

\bibitem[{Wang et~al.(2022{\natexlab{b}})Wang, Zhang, Ebrahimi, Sun, Zhang, Lee, Ren, Su, Perot, Dy et~al.}]{dualprompt}
Wang, Z.; Zhang, Z.; Ebrahimi, S.; Sun, R.; Zhang, H.; Lee, C.-Y.; Ren, X.; Su, G.; Perot, V.; Dy, J.; et~al. 2022{\natexlab{b}}.
\newblock Dualprompt: Complementary prompting for rehearsal-free continual learning.
\newblock In \emph{European Conference on Computer Vision}, 631--648. Springer.

\bibitem[{Wang et~al.(2022{\natexlab{c}})Wang, Zhang, Lee, Zhang, Sun, Ren, Su, Perot, Dy, and Pfister}]{L2P}
Wang, Z.; Zhang, Z.; Lee, C.-Y.; Zhang, H.; Sun, R.; Ren, X.; Su, G.; Perot, V.; Dy, J.; and Pfister, T. 2022{\natexlab{c}}.
\newblock Learning to prompt for continual learning.
\newblock In \emph{Proceedings of the IEEE/CVF Conference on Computer Vision and Pattern Recognition}, 139--149.

\bibitem[{Wu et~al.(2019)Wu, Chen, Wang, Ye, Liu, Guo, and Fu}]{BiC}
Wu, Y.; Chen, Y.; Wang, L.; Ye, Y.; Liu, Z.; Guo, Y.; and Fu, Y. 2019.
\newblock Large scale incremental learning.
\newblock In \emph{Proceedings of the IEEE/CVF conference on computer vision and pattern recognition}, 374--382.

\bibitem[{Xiao, Rasul, and Vollgraf(2017)}]{fashionmnist}
Xiao, H.; Rasul, K.; and Vollgraf, R. 2017.
\newblock Fashion-mnist: a novel image dataset for benchmarking machine learning algorithms.
\newblock \emph{arXiv preprint arXiv:1708.07747}.

\bibitem[{Xie, Yan, and He(2022)}]{UCL3}
Xie, J.; Yan, S.; and He, X. 2022.
\newblock General incremental learning with domain-aware categorical representations.
\newblock In \emph{Proceedings of the IEEE/CVF Conference on Computer Vision and Pattern Recognition}, 14351--14360.

\bibitem[{Yao and Miller(2015)}]{tinyimagenet}
Yao, L.; and Miller, J. 2015.
\newblock Tiny imagenet classification with convolutional neural networks.
\newblock \emph{CS 231N}, 2(5): 8.

\bibitem[{Zhou et~al.(2022)Zhou, Wang, Ye, and Zhan}]{memo}
Zhou, D.-W.; Wang, Q.-W.; Ye, H.-J.; and Zhan, D.-C. 2022.
\newblock A model or 603 exemplars: Towards memory-efficient class-incremental learning.
\newblock \emph{arXiv preprint arXiv:2205.13218}.

\end{thebibliography}

\newpage
\newpage
\appendix
{\bf{\huge Appendix}}

In the supplemental document, we present:
\begin{itemize}
    \item Evaluation Details
    \item Details of Empirical Study
    \item Details of \methodname implementation, evaluation benchmarks, and additional results
    \item Limitations and Future Work
\end{itemize}

\section{Evaluation Metrics Details}
\subsection{Metrics of class incremental learning}
\label{sec: Appendix: CIL metrics}
Based on the survey~\cite{zhangxingxing}, we evaluate the continual model mainly from two aspects: the overall \textbf{A}verage \textbf{A}ccuracy $AA$ and \textbf{F}orgetting \textbf{M}easure $FM$.

Let $a_{k,j}\in[0,100]$ denote the classification accuracy evaluated on the test dataset of the $j$-th task after incremental training of the $k$-th task($1\leq j\leq k$). The average accuracy at the $k$-th task is defined as:
\begin{align}
    AA_k=\frac {1}{k}\sum_{j=1}^{k} a_{k, j}
\end{align}
Furtherly, the overall average performance $AA$ after all $T$ tasks is then defined as:
\begin{align}
    AA=\frac 1T \sum_{k=1}^{T} AA_k
\end{align}

For a single task, the forgetting \(f\) is calculated by the difference between its maximum accuracy achieved in the past and its current accuracy after the $k$-th task:
\begin{align}
    f_{j, k}=\max _{i \in\{1, \ldots, k-1\}}\left(a_{i, j}-a_{k, j}\right), \forall j<k.
\end{align}
Then $FM$ after all \(T\) task is the average forgetting of all old tasks:
\begin{align}
    \mathrm{FM}=\frac{1}{T-1} \sum_{j=1}^{T-1} f_{j, T-1}.
\end{align}

Moreover, to reflect the model performance when learning a new task, we design the \textbf{I}nitial \textbf{A}ccuracy $IA$, which is the average accuracy of each task after its initial training:
\begin{align}
    IA=\frac{1}{T}\sum_{i=1}^Ta_{i,i}
\end{align}

\subsection{Metrics $PIV$ and $PFTS$}
\label{sec: Appendix: CILD metrics}
We decouple the model $f_\theta$ into two parts: the feature extractor $\mathbf{F}$ parameterized by $\theta_\mathcal{F}$ and the classifier $\mathbf{W}$ parameterized by $\theta_\mathcal{W}$. Since we aim to investigate parameters related to features, we represent the parameter update between task \(t\) and task \(t-1\) as \(\Delta\theta_t=\theta_{\mathcal{F}_t}-\theta_{\mathcal{F}_{t-1}}\). For each task \(t\), we use \(\mathcal{H}_t=\left\{i||\left(\Delta \theta_t\right)_i \mid>\delta\right\}\) (\(\delta\) is the threshold which is set to be the upper quartile value of all parameters' update) to denote the set of indices corresponding to the high-magnitude updates in \(\Delta\theta_t\). For each pair of tasks \(t\) and \(t'\), we use \(J(t,\ t')\) to measure the overlap in parameters they update. Specifically, the interference value \(IS\) is quantified using the Jaccard index\cite{jaccard1901etude}, which is a measure of similarity between finite sets:
\begin{align}
    IS=J\left(t, t^{\prime}\right)=\frac{\left|\mathcal{H}_t \cap \mathcal{H}_{t^{\prime}}\right|}{\left|\mathcal{H}_t \cup \mathcal{H}_{t^{\prime}}\right|}.
\end{align}
When the interference between task \(t\) and \(t^{\prime}\) is small, the \(IS\) will be correspondingly lower. We then compute the pairwise forward transfer score \(FTS\) as below:
\begin{align}
    FTS\left(t, t^{\prime}\right)=J\left(t, t^{\prime}\right) \times\left(\frac{\left\|\Delta \theta_t\right\|_2+\left\|\Delta \theta_{t^{\prime}}\right\|_2}{2}\right).
\end{align}
This score captures both the overlap and average update magnitude, providing a measurement of knowledge learned by parameters in task \(t\). The higher \(FTS\) is, the more the model learns at stage \(t\). Consequently, \(FTS\) can be regarded as the learning plasticity of the model.
Then the overall parameter interference value \(PIV\) and overall parameter forward transfer \(PFTS\)  is represented as:
\begin{align}
    \mathrm{PIV}=\frac{1}{T-1} \sum_{t=2}^{T} IS\left(t, t-1\right)\\
    \mathrm{PFTS}=\frac{1}{T-1} \sum_{t=2}^{T} FTS\left(t, t-1\right)
\end{align}

\section{Details of Empirical Study}
\begin{table*}[ht]
    \centering
    
    \begin{tabular}{cc|ccl|ccl}
    \toprule
                             &                            & \multicolumn{3}{c|}{CIFAR-100}                                                                                              & \multicolumn{3}{c}{DomainNet}                                                                                               \\
    \multirow{-2}{*}{Method} & \multirow{-2}{*}{scenario} & $AA\uparrow$                           & $FM\downarrow$                                 & backbone                          & $AA\uparrow$                           & $FM\downarrow$                                 & backbone                          \\ \midrule
    iCaRL                    & CIL                        & \textbf{64.15}                         & 46.23                                          & ViT-B/16                          & 52.51                                  & 34.19                                          & ViT-B/16                          \\
    iCaRL                    & CILD                       & \cellcolor[HTML]{C0C0C0}63.69          & \cellcolor[HTML]{C0C0C0}\textbf{22.00(-24.23)} & \cellcolor[HTML]{C0C0C0}ViT-B/16  & \cellcolor[HTML]{C0C0C0}\textbf{56.15} & \cellcolor[HTML]{C0C0C0}\textbf{18.05(-16.14)} & \cellcolor[HTML]{C0C0C0}ViT-B/16  \\ \midrule
    BiC                      & CIL                        & 68.55                                  & 36.60                                          & ViT-B/16                          & \textbf{59.22}                         & 34.25                                          & ViT-B/16                          \\ 
    BiC                      & CILD                       & \cellcolor[HTML]{C0C0C0}\textbf{69.03} & \cellcolor[HTML]{C0C0C0}\textbf{12.11(-24.49)} & \cellcolor[HTML]{C0C0C0}ViT-B/16  & \cellcolor[HTML]{C0C0C0}57.98          & \cellcolor[HTML]{C0C0C0}\textbf{16.23(-18.02)} & \cellcolor[HTML]{C0C0C0}ViT-B/16  \\ \midrule
    MEMO                     & CIL                        & 75.63                                  & 25.44                                          & ResNet-18                         & 76.57                                  & 16.35                                          & ResNet-32                         \\
    MEMO                     & CILD                       & \cellcolor[HTML]{C0C0C0}\textbf{76.89} & \cellcolor[HTML]{C0C0C0}\textbf{16.10(-9.34)}  & \cellcolor[HTML]{C0C0C0}ResNet-18 & \cellcolor[HTML]{C0C0C0}\textbf{78.93} & \cellcolor[HTML]{C0C0C0}\textbf{6.87(-9.48)}   & \cellcolor[HTML]{C0C0C0}ResNet-32 \\ \midrule
    LwF                      & CIL                        & 53.23                                  & 43.53                                          & ViT-B/16                          & \textbf{48.08}                         & 42.16                                          & ViT-B/16                          \\
    LwF                      & CILD                       & \cellcolor[HTML]{C0C0C0}\textbf{54.07} & \cellcolor[HTML]{C0C0C0}\textbf{16.26(-27.27)} & \cellcolor[HTML]{C0C0C0}ViT-B/16  & \cellcolor[HTML]{C0C0C0}45.00          & \cellcolor[HTML]{C0C0C0}\textbf{21.84(-20.32)} & \cellcolor[HTML]{C0C0C0}ViT-B/16  \\ \midrule
    DER                      & CIL                        & 64.64                                  & 45.15                                          & ResNet-18                         & 72.18                                  & 24.63                                          & ResNet-32                         \\
    DER                      & CILD                       & \cellcolor[HTML]{C0C0C0}\textbf{66.37} & \cellcolor[HTML]{C0C0C0}\textbf{2.25(-42.90)}  & \cellcolor[HTML]{C0C0C0}ResNet-18 & \cellcolor[HTML]{C0C0C0}\textbf{73.49} & \cellcolor[HTML]{C0C0C0}\textbf{13.50(-11.13)} & \cellcolor[HTML]{C0C0C0}ResNet-32 \\
    \bottomrule
    \end{tabular}
    \caption{Comparative analysis on accuracy($AA$) and forgetting($FM$) of CIL models under CIL and CILD scenarios after alternating their backbones.}
    \label{tab: CIL vs CILD backbone}
\end{table*}
\subsection{Details of the construction of CIL and CILD scenarios}
\label{sec: Appendix: Details of Empirical Study: construction of CILD}
\subsubsection{Synthesizing images of new domains}. We manually add domain shift to the original CIFAR-100 dataset using a style transfer GAN named Avatar-Net~\cite{avatarnet} to synthesize images of new domains, resulting in a new dataset DomainCIFAR-100. 
In the official implementation of AvatarNet, the author provides 5 styles: \textit{brush strokes}, \textit{lamuse}, \textit{plum flower}, \textit{woman in peasant dress}, and \textit{candy}. For the original CIFAR-100 which contains 60000 images and 100 labels, we feed it into AvatarNet and get $5\times 60000$ images of 5 domains. We split the class space [0,99] randomly into 6 dis-joint subspaces $S_i$, with the first subspace containing 50 classes and the others containing 10 classes respectively. At the same time, we randomly design a 6-domain order $\mathcal{P}$ such as \{real, brush strokes, lamuse, plum flower, woman in peasant dress, candy\}, where \textit{real} means the original style. Then as shown in Fig. ~\ref{fig: Illustration of CILD}, we can construct a DomainCIFAR-100 CILD scenario with 6 tasks, each task containing images of one style. The corresponding CIL scenario shares the same subspaces with that of CILD, but all tasks are in the same domain (We choose the first domain in $\mathcal{P}$ of CILD as the only domain of CIL). By this design, we can simulate the situation where each task is introduced with a new domain while maintaining its image structure to the most extent.

\subsubsection{Splitting DomainNet}
DomainNet~\cite{domainnet} is a popular dataset in transfer learning, spanning over 6 domains: \textit{clipart}, \textit{infograph}, \textit{painting}, \textit{quickdraw}, \textit{real}, and \textit{sketch}, each domain has 345 classes. Similarly, we randomly split the class space [0,344] into 5 dis-joint subspaces $S_i$, each subspace has $345\div 5=69$ classes. We randomly design a 6-domain order $\mathcal{P}$ such as \{real, infograph, painting, quickdraw, clipart, and sketch\}. The same as we construct DomainCIFAR-100, we can then construct CILD and CIL scenarios of DomainNet. Compared to DomainCIFAR-100, CILD scenario of DomainNet is more similar to real-world application where both styles and image structures may change significantly in different tasks.

\subsection{Details of empirical study implementation}
The backbones of used baselines in section ~\ref{sec: Empirical Study: observation} are shown in Tab. ~\ref{tab: appendix: backbones list}. 
\begin{table}[ht]
    \centering
    
    \begin{tabular}{ccc}
    \toprule
    \multirow{2}{*}{Method} & \multicolumn{2}{c}{Backbone} \\
                            & CIFAR-100     & DomainNet    \\ \midrule
    iCaRL                   & ResNet-18     & ResNet-32    \\
    BiC                     & ResNet-18     & ResNet-32    \\
    MEMO                    & ViT-B/16      & ViT-B/16     \\
    LwF                     & ResNet-18     & ResNet-32    \\
    DER                     & ViT-B/16      & ViT-B/16     \\
    L2P                     & ViT-B/16      & ViT-B/16    \\
    \bottomrule
    \end{tabular}
    \caption{Backbones of baselines in Empirical Study. We choose these backbones based on the default config of the PILOT repository~\cite{pilot1}.}
    \label{tab: appendix: backbones list}
\end{table}
We train all models from scratch with random initialization. The implementation code and training strategy are the same with that of section ~\ref{sec: experiment: implementation}.

\subsection{More details of observation}
\label{sec: Appendix: Details of Empirical Study: more observations}

\subsubsection{Observation on different backbones}
To validate the generalizability of our observation, we change the backbone and conduct exactly the same experiments as in section ~\ref{sec: Empirical Study: observation}, the result is shown in Tab. ~\ref{tab: CIL vs CILD backbone}. We can observe that no matter the backbone type, models under CILD scenario demonstrate better performance than CIL.

\subsubsection{Observation on pre-trained models}
For ViT-based models, we use the pre-trained model provided by PILOT~\cite{pilot1} to initialize our model, and for ResNet-based models, we load the pre-trained resnet weights from pytorch. We conduct the same empirical study as shown in Tab. ~\ref{tab: CIL vs CILD pretrain}. All models are trained 20 epochs for each task using the backbone type as in Tab. ~\ref{tab: appendix: backbones list}. We can get the same observation under this circumstance.

\subsubsection{Observation on first task}
\label{sec: Appendix: observation of the first task}
\begin{figure}[h]
  \centering
  \includegraphics[width=1.0\columnwidth]{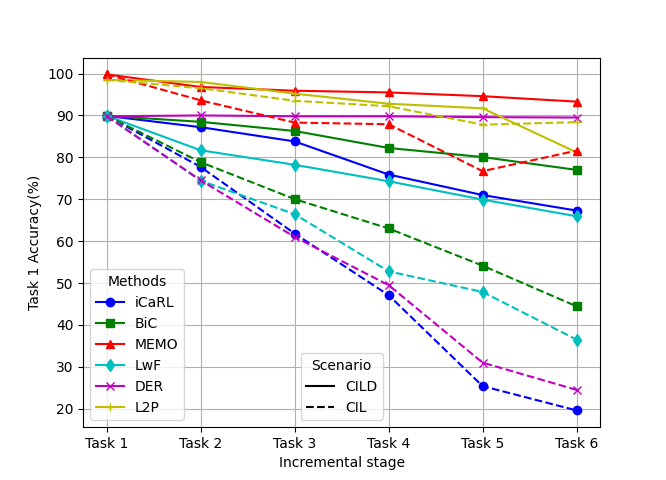}
    \caption{Comparison of accuracy curves of the first task on CIFAR-100. Most continual methods under CILD demonstrate a fantastic low forgetting rate of the first task.}
    \label{fig: task0_acc}
\end{figure}

As shown in Fig. ~\ref{fig: task0_acc}, if we take a close look at just the first task performance during the whole training process, we can observe that models suffer from much less forgetting under CILD.

\begin{table*}[ht]
\centering

\begin{tabular}{cc|ccl|ccl}
\toprule
                         &                            & \multicolumn{2}{c}{CIFAR-100}                                                           &                                   & \multicolumn{2}{c}{DomainNet}                                                           &                                   \\
\multirow{-2}{*}{Method} & \multirow{-2}{*}{scenario} & $AA\uparrow$                           & $FM\downarrow$                                 & \multirow{-2}{*}{Backbone}        & $AA\uparrow$                           & $FM\downarrow$                                 & \multirow{-2}{*}{Backbone}        \\ \midrule
iCaRL                    & CIL                        & 81.51                                  & 25.06                                          & ResNet-18                         & 83.62                                  & 23.60                                          & ResNet-32                         \\
iCaRL                    & CILD                       & \cellcolor[HTML]{C0C0C0}\textbf{83.02} & \cellcolor[HTML]{C0C0C0}\textbf{8.57(-24.23)}  & \cellcolor[HTML]{C0C0C0}ResNet-18 & \cellcolor[HTML]{C0C0C0}\textbf{84.29} & \cellcolor[HTML]{C0C0C0}\textbf{9.34(-14.26)}  & \cellcolor[HTML]{C0C0C0}ResNet-32 \\ \midrule
BiC                      & CIL                        & 83.24                                  & 29.60                                          & ResNet-18                         & \textbf{79.26}                         & 29.84                                          & ResNet-32                         \\
BiC                      & CILD                       & \cellcolor[HTML]{C0C0C0}\textbf{85.00} & \cellcolor[HTML]{C0C0C0}\textbf{7.67(-21.93)}  & \cellcolor[HTML]{C0C0C0}ResNet-18 & \cellcolor[HTML]{C0C0C0}76.21          & \cellcolor[HTML]{C0C0C0}\textbf{10.25(-19.59)} & \cellcolor[HTML]{C0C0C0}ResNet-32 \\ \midrule
MEMO                     & CIL                        & \textbf{90.38}                         & 13.29                                          & ViT-B/16                          & \textbf{92.06}                         & 8.64                                           & ViT-B/16                          \\
MEMO                     & CILD                       & \cellcolor[HTML]{C0C0C0}90.17          & \cellcolor[HTML]{C0C0C0}\textbf{7.01(-6.28)}   & \cellcolor[HTML]{C0C0C0}ViT-B/16  & \cellcolor[HTML]{C0C0C0}88.29          & \cellcolor[HTML]{C0C0C0}\textbf{5.49(-3.15)}   & \cellcolor[HTML]{C0C0C0}ViT-B/16  \\ \midrule
LwF                      & CIL                        & \textbf{74.68}                         & 26.89                                          & ResNet-18                         & \textbf{70.64}                         & 29.16                                          & ResNet-32                         \\
LwF                      & CILD                       & \cellcolor[HTML]{C0C0C0}72.03          & \cellcolor[HTML]{C0C0C0}\textbf{14.35(-12.54)} & \cellcolor[HTML]{C0C0C0}ResNet-18 & \cellcolor[HTML]{C0C0C0}68.33          & \cellcolor[HTML]{C0C0C0}\textbf{16.58(-12.58)} & \cellcolor[HTML]{C0C0C0}ResNet-32 \\ \midrule
DER                      & CIL                        & \textbf{87.04}                         & 16.29                                          & ViT-B/16                          & \textbf{89.18}                         & 20.98                                          & ViT-B/16                          \\
DER                      & CILD                       & \cellcolor[HTML]{C0C0C0}86.80          & \cellcolor[HTML]{C0C0C0}\textbf{0.40(-42.90)}  & \cellcolor[HTML]{C0C0C0}ViT-B/16  & \cellcolor[HTML]{C0C0C0}\textbf{88.22} & \cellcolor[HTML]{C0C0C0}\textbf{8.42(-12.56)}  & \cellcolor[HTML]{C0C0C0}ViT-B/16 \\ \bottomrule
\end{tabular}
\caption{Comparative analysis on accuracy($AA$) and forgetting($FM$) of pre-trained CIL models under CIL scenario and CILD scenario. ViT-based models load all weights from PILOT~\cite{pilot1}, while ResNet-based models load backbone weight from pytorch.}
\label{tab: CIL vs CILD pretrain}
\end{table*}

\subsubsection{Observation on different domain orders}
The results above in the empirical study are all achieved under the domain order where the first domain is real. We alter the first domain and conduct more experiments. For CIFAR-100, the CILD domain order is: 

\{\textit{brush strokes}$\rightarrow$\textit{candy}$\rightarrow$\textit{real}$\rightarrow$\textit{woman in peasant dress}$\rightarrow$\textit{lamuse}$\rightarrow$\textit{plum flower}\}, 

while the CIL has the only domain \textit{brush strokes}. For DomainNet, the CILD domain order is: 

\{\textit{quickdraw}$\rightarrow$\textit{painting}$\rightarrow$\textit{clipart}$\rightarrow$\textit{real}$\rightarrow$\textit{infograph}$\rightarrow$\textit{sketch}\}, 

while the CIL has the only domain \textit{quickdraw}.

We can the same observation in Tab. ~\ref{tab: CIL vs CILD domain order} that these models demonstrate a fantastic lower forgetting rate under the CILD scenario. Moreover, the forgetting rate under CILD scenario in this domain order is comparable or even lower than in the original order. In contrast, most models suffer from worse forgetting under CIL scenario in this domain order.
\begin{table}[ht]
    \centering
    \setlength{\tabcolsep}{1mm}
    \small{
    \begin{tabular}{cc|cc|cc}
    \toprule
                             &                            & \multicolumn{2}{c|}{CIFAR-100}                                                          & \multicolumn{2}{c}{DomainNet}                                                           \\
    \multirow{-2}{*}{Method} & \multirow{-2}{*}{scenario} & $AA\uparrow$                           & $FM\downarrow$                                 & $AA\uparrow$                           & $FM\downarrow$                                 \\ \midrule
    iCaRL                    & CIL                        & 52.34                                  & 53.67                                          & 53.09                                  & 55.60                                          \\
    iCaRL                    & CILD                       & \cellcolor[HTML]{C0C0C0}\textbf{55.15} & \cellcolor[HTML]{C0C0C0}\textbf{23.87(-29.80)} & \cellcolor[HTML]{C0C0C0}\textbf{56.21} & \cellcolor[HTML]{C0C0C0}\textbf{19.63(-35.97)} \\ \midrule
    BiC                      & CIL                        & 64.98                                  & 43.16                                          & \textbf{57.30}                         & 46.14                                          \\
    BiC                      & CILD                       & \cellcolor[HTML]{C0C0C0}\textbf{70.19} & \cellcolor[HTML]{C0C0C0}\textbf{15.30(-27.86)} & \cellcolor[HTML]{C0C0C0}56.77          & \cellcolor[HTML]{C0C0C0}\textbf{15.27(-30.87)} \\ \midrule
    MEMO                     & CIL                        & \textbf{84.29}                         & 21.44                                          & \textbf{83.66}                         & 16.84                                          \\
    MEMO                     & CILD                       & \cellcolor[HTML]{C0C0C0}81.67          & \cellcolor[HTML]{C0C0C0}\textbf{9.56(-11.88)}  & \cellcolor[HTML]{C0C0C0}79.25          & \cellcolor[HTML]{C0C0C0}\textbf{7.98(-8.86)}   \\ \midrule
    LwF                      & CIL                        & \textbf{46.28}                         & 53.10                                          & \textbf{44.35}                         & 51.20                                          \\
    LwF                      & CILD                       & \cellcolor[HTML]{C0C0C0}43.76          & \cellcolor[HTML]{C0C0C0}\textbf{26.54(-26.56)} & \cellcolor[HTML]{C0C0C0}42.43          & \cellcolor[HTML]{C0C0C0}\textbf{24.22(-26.98)} \\ \midrule
    DER                      & CIL                        & 58.32                                  & 41.26                                          & \textbf{62.19}                         & 34.18                                          \\
    DER                      & CILD                       & \cellcolor[HTML]{C0C0C0}\textbf{62.87} & \cellcolor[HTML]{C0C0C0}\textbf{1.3(-39.96)}   & \cellcolor[HTML]{C0C0C0}60.36          & \cellcolor[HTML]{C0C0C0}\textbf{11.75(-22.43)} \\ \midrule
    L2P                      & CIL                        & \textbf{77.41}                         & \textbf{11.54}                                 & \textbf{74.21}                         & \textbf{13.57}                                 \\
    L2P                      & CILD                       & \cellcolor[HTML]{C0C0C0}56.25          & \cellcolor[HTML]{C0C0C0}23.80(+12.26)          & \cellcolor[HTML]{C0C0C0}51.07          & \cellcolor[HTML]{C0C0C0}26.70(+13.13)        \\ \bottomrule 
    \end{tabular}
    }
    \caption{Comparative analysis on accuracy($AA$) and forgetting($FM$) of CIL models under CIL and CILD scenarios after alternating the domain order.}
    \label{tab: CIL vs CILD domain order}
\end{table}

\subsubsection{Dive into the ``forgetting" of CILD}
Forgetting is complex and may not be perfectly reflected by the accuracy drop. In CILD, there is possibility that the model \textbf{does not indeed ``forget less" its inner knowledge} about classifying seen categories, but rather \textbf{more able to ``recall" its memory}. That is to say, there is a chance that adding domain shift may help model identify the task boundary, hence help model ``recall" its memory. 

\begin{table}[h]
    \centering
    \small{
        \begin{tabular}{cc|cc}
        \toprule
        Method                 & Scenario & $TIA$   & $ITA-first$ \\ \midrule
        \multirow{2}{*}{iCaRL} & CIL      & 50.14 & 36.22     \\
                               & CILD     & \textbf{84.29} & \textbf{79.06}     \\ \midrule
        \multirow{2}{*}{BiC}   & CIL      & 51.06 & 37.64     \\
                               & CILD     & \textbf{86.22} & \textbf{81.55}     \\ \midrule
        \multirow{2}{*}{MEMO}  & CIL      & 84.16 & 85.12     \\
                               & CILD     & \textbf{91.06} & \textbf{92.99}     \\ \midrule
        \multirow{2}{*}{LwF}   & CIL      & 48.98 & 42.13     \\
                               & CILD     & \textbf{85.35} & \textbf{83.83}     \\ \midrule
        \multirow{2}{*}{DER}   & CIL      & 82.69 & 35.12     \\
                               & CILD     & \textbf{98.67} & \textbf{96.75}     \\ \midrule
        \multirow{2}{*}{L2P}   & CIL      & \textbf{92.65} & \textbf{90.46}     \\
                               & CILD     & 81.28 & 83.69    \\ \bottomrule
        \end{tabular}
    }
    \caption{Results of Task-Inference Accuracy($TIA$) and (first)Intra-Task Accuracy($ITA$) on CIFAR-100. Higher $TIA$ means model is more able to identify task ID, thus being more able to recall its corresponding knowledge. Higher $ITA$ means the model remembers more knowledge about certain classes.}
    \label{tab: TIA and ITA}
\end{table}

To investigate this, we employ two metrics: \textbf{T}ask-\textbf{I}nference \textbf{A}ccuracy($TIA$) to evaluate the ability to identify task ID(ability to recall memory), and \textbf{I}ntra-\textbf{T}ask \textbf{A}ccuracy($ITA$) to evaluate the inner knowledge of model. For $TIA$, after training of all tasks, we give test sample to the model and see which label space(i.e. inferenced task ID) does this predicted class belongs to. For $ITA$, we isolate the weight of the classifier and treat each of them as the exclusive classifier for each task. We manually feed the sample feature to its corresponding classifier and get its $ITA$ in that classifier. We only compare the $ITA$ of the first task since they share the same test set. From Tab. ~\ref{tab: TIA and ITA}, we can see that domain shift in CILD indeed helps model identify the task boundary. Moreover, $ITA$ in CIL is much lower, indicating model does forget what it learned. So, the performance under CILD is not only caused by a clearer task boundary, but also due to forgetting less knowledge.

\section{Details of \methodname implementation, evaluation benchmarks, and additional results}
\subsection{Datasets Details}
\label{sec: Appendix: dataset}
\begin{itemize}
    \item CIFAR100~\cite{cifar100} is a classic classification dataset composed of 100 categories, each having 500 training images and 100 test images of resolution 32$\times$32 pixels. We follow the setting of previous works\cite{AA, icarl} and split the CIFAR100 randomly and evenly into 10 tasks, each tasks having 10 classes respectively.
    \item Fashion-MNIST~\cite{fashionmnist} is a dataset of Zalando's article images—consisting of a training set of 60,000 examples and a test set of 10,000 examples. Each example is a 28$\times$28 grayscale image, associated with a label from 10 classes. The Fashion-MNIST dataset offers a more challenging benchmark than the traditional MNIST due to its more complex and varied patterns, which better mimic real-world data. This increased complexity helps develop and test more robust machine learning models. We split Fashion-MNIST randomly and evenly into 5 tasks, each task containing 2 classes.
    \item Due to resource limits, we employ Tiny-ImageNet~\cite{tinyimagenet} instead of the full ImageNet~\cite{imagenet} as one of the benchmarks. Tiny-ImageNet is a simplified version of the larger ImageNet, covering 200 classes, and each contains 500 training images, 50 validation images, and 50 test images. The images are downscaled to 64x64 pixels, compared to the higher resolutions in the original ImageNet. We split Tiny-ImageNet into 10 tasks randomly and evenly, each task containing 20 classes.
\end{itemize}
\begin{figure*}
    \centering
    \includegraphics[width=1.8\columnwidth]{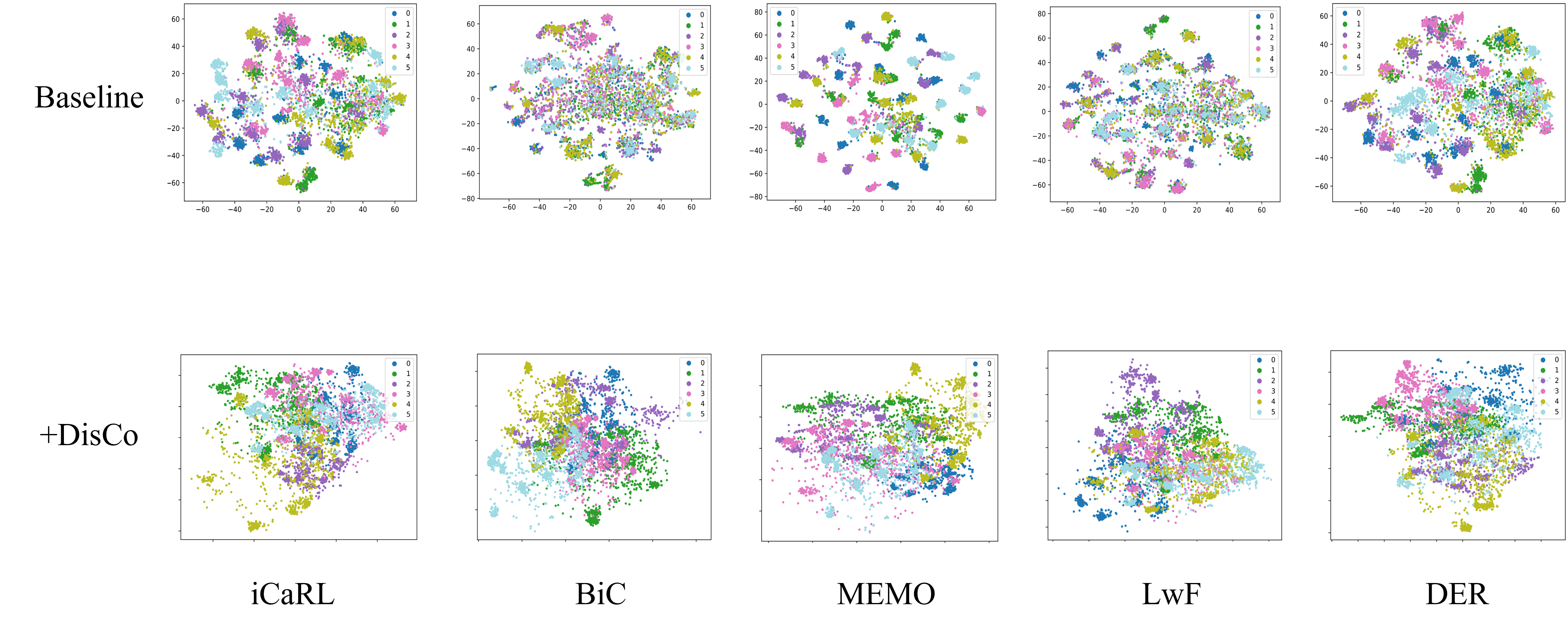}
    \caption{t-SNE visualization of features on CIFAR-100. The top row denotes features extracted by different baseline continual methods and the bottom row denotes features of incorporating \methodname into them. Data points from the same task are marked using the same color.}
    \label{fig: appendix: main_tsne}
\end{figure*}
\subsection{Baseline Details}
\label{sec: Appendix: baseline}
\begin{itemize}
    \item iCaRL~\cite{icarl} is a rehearsal-based class incremental learning model, proposing a novel method to choose buffer samples to replay and using KD loss to regulate the logits of old and new models.

    \item MEMO~\cite{memo} proposes decoupling the backbone at middle layers into shallow and deep layers and expands deep layers for each task depending on the limited memory buffer.

    \item BiC~\cite{BiC}, or \textbf{Bi}as \textbf{C}orrection, is a rehearsal-based model designed to enhance continual learning models by addressing the issue of biased representation towards newer classes. It does this by introducing a bias correction layer at the end of the learning model, which adjusts the decision boundaries in favor of previously learned tasks.

    \item LwF~\cite{lwf} is a regularization-based model that uses knowledge distillation to align the outputs of old and new models without a memory buffer.

    \item DER~\cite{DER} adapt the dynamic structure from Task-Incremental Learning to Class-Incremental Learning, which freezes the old feature model to preserve old knowledge while creating a new trainable feature extractor concatenating their features together to adapt the model to new tasks. Like rehearsal-based methods, DER requires a memory buffer to store old task samples.

    \item L2P~\cite{L2P} is a recently popular prompt-based method, which leverages the frozen pre-trained vision transformer~\cite{vit} with trainable prompts to adapt the model to continual scenarios.

\end{itemize}
The backbones used by them are the same as those in empirical study, as listed in Tab. ~\ref{tab: appendix: backbones list}. For ViT-based methods, we up-sample the original images to $224\times 224$ using bicubic interpolation. 

\subsection{Incorporating \methodname into prompt-based methods}
Prompt-based methods like L2P~\cite{L2P} leverage a frozen pre-trained vision transformer~\cite{vit} with a trainable prompt pool to adapt the model to continual scenarios. The prompt pool $\mathcal{PP}$ is composed of $M$ key-value pairs $\mathcal{PP}={\{k_1,v_1\}, \{k_2,v_2\}...\{k_M, v_M\}}$, where $v_i\in \mathbb{R}^{L_p\times d}$ is a prompt with token length $L_p$ and $k_i\in \mathbb{R}^d$ is the key attached to it. Given an image input $x$, L2P first uses a frozen pre-trained vision transformer $\phi_0$ to extract the CLS token $q\in \mathbb{R}^d$. Then L2Ps use $q$ as the query to lookup top-N nearest keys $K=\{k_1, k_2,...k_N\}$ and their corresponding prompts $V=\{v_1,v_2,...v_N\}$. The selected prompts $V\in \mathbb{R}^{(N\times L_p)\times d}$ are appended before the patch embeddings of $x$ and feed into $\phi_0$ for classification.

\label{sec: Appendix: incorporating into l2p}
To incorporate \methodname into it, the key idea is the same as that of incorporating \methodname into other methods: find prototype $P$ of current task $t$ and push current features towards $P$ while keeping them away from previous prototypes. However, directly contrasting CLS features of L2P may lead to even worse forgetting as suggested in ~\ref{sec: Empirical Study: observation}. \textbf{Instead, we contrast the selected keys $K$.}

For a mini-batch of batchsize $N_b$, we first count the frequency of each key being selected. We treat the most frequently selected $N$ keys as ``features of current task". For \methodname-I, we use their average mean as batch-wise prototype $p_i$.  We impose the same task-level regularization on these keys as in Eq. ~\ref{eq: tcon}, with $p_i$ being the positive sample and prototypes of previous tasks as negative samples.

\subsection{Additional results}
\label{sec: Appendix: additional results}
\subsubsection{t-SNE visualization of incorporating DisCo}
We use t-SNE to visualize the features of incorporating DisCo into baseline models, as shown in Fig. ~\ref{fig: appendix: main_tsne}.

We can see that incorporating \methodname yields relatively more distinguishable features, with comparatively more isolated feature distribution for each task.

\subsubsection{Sensitiveness to blurry task boundary}
DisCo relies on prototypes, and a blurry task boundary may influence performance. So we evaluate Disco on CUB200~\cite{cub}, which is a fine-grained classification dataset covering 200 types of birds. As in Tab. ~\ref{tab:cub result}, DisCo handles this challenge well, reducing forgetting and improving average accuracy. However, it’s worth noting that the $AA$ improvement and $FM$ reduction on CUB200 are not that significant compared to the three datasets in the main study Tab. ~\ref{tab: Main Result}. This means blurry task boundaries do have some minor effect on DisCo, which will be a research point for future works.
\begin{table}[t]
\centering
    \begin{tabular}{c|cc}
    \toprule
    Method        & $AA\uparrow$    & $FM\downarrow$    \\ \midrule
    iCaRL         & 52.06 & 41.95 \\
    iCaRL+DisCo-I & \textbf{52.43} & \textbf{36.16} \\ \midrule
    BiC           & 53.98 & 36.08 \\
    BiC+DisCo-I   & \textbf{54.32} & \textbf{33.39} \\ \midrule
    LwF           & 44.49 & 46.44 \\
    LwF+DisCo-I   & \textbf{46.84} & \textbf{38.99} \\ \midrule
    DER           & 52.73 & 36.32 \\
    DER+DisCo-I   & \textbf{54.09} & \textbf{34.21} \\ \midrule
    L2P           & \textbf{71.50} & \textbf{8.58}  \\
    L2P+DisCo-T   & 71.29 & 8.97 \\ \bottomrule
    \end{tabular}
\caption{Result of proposed DisCo on fine-grained dataset CUB200(10 classes per task). DisCo succeeds in handling this challenge but its performance boost is restricted due to the blurry task boundary.}
\label{tab:cub result}
\end{table}

\subsection{Ablation study on loss weight $\lambda$}
\label{sec: Appendix: ablation study on lambda}
In Tab. ~\ref{tab: ablation study on lambda iCaRL}-Tab. ~\ref{tab: ablation study on lambda L2P}, we analyze the influence of the weight balance $\lambda_{tcon}$, $\lambda_{ccon}$ and $\lambda_{ccd}$ of Eq. ~\ref{eq: loss} on CIFAR-100. 

We can see that \methodname achieves a consistently good performance when $\lambda_{tcon}$ and $\lambda_{ccon}$ are set to 0.5 and $\lambda_{ccd}=1$. When we amplify $\lambda_{tcon}$, which means we impose stricter regularization on task-level distributions, and this will make $FM$ drop but $AA$ drops too. This is consistent with the ablation study of different modules in section ~\ref{sec: ablation study on modules}.

\begin{table}[h]
    \centering
    
    \setlength{\tabcolsep}{1mm}
        \begin{tabular}{cccccc}
        \toprule
        Method                         & $\lambda_{tcon}$ & $\lambda_{ccon}$ & $\lambda_{ccd}$ & $AA\uparrow$                        & $FM\downarrow$    \\ \midrule
        \multirow{4}{*}{iCaRL DisCo-I} & 0.5              & 0.5              & 1.0             & \textbf{70.11} & 33.96 \\
                                       & 1.0              & 0.5              & 1.0             & 68.56 & \textbf{32.47} \\
                                       & 0.5              & 1.0              & 1.0             & 67.44 & 34.80 \\
                                       & 1.0              & 1.0              & 1.0             & 69.16 & 33.87 \\ 
                                       & 1.0              & 1.0              & 0.5             & \multicolumn{1}{l}{68.64} & \multicolumn{1}{l}{36.47}\\
                                       
                                       \bottomrule
        \end{tabular}%
    \caption{Evaluation of other possible values of loss weight $\lambda$ on iCaRL.}
    \label{tab: ablation study on lambda iCaRL}
        
\end{table}

\begin{table}[hb]
    \centering
    
    \begin{tabular}{cccccc}
    \toprule
    Method                       & $\lambda_{tcon}$ & $\lambda_{ccon}$ & $\lambda_{ccd}$ & $AA\uparrow$                        & $FM\downarrow$                        \\ \midrule
    \multirow{5}{*}{BiC DisCo-I} & 0.5              & 0.5              & 1.0             & \textbf{69.89}            & 28.54                     \\
                                 & 1.0              & 0.5              & 1.0             & 67.21                     & \textbf{27.34}            \\
                                 & 0.5              & 1.0              & 1.0             & 66.24                     & 27.67                     \\
                                 & 1.0              & 1.0              & 1.0             & 62.13                     & 30.16                     \\
                                 & 1.0              & 1.0              & 0.5             & \multicolumn{1}{l}{61.76} & \multicolumn{1}{l}{31.01} \\ \bottomrule
    \end{tabular}
    \caption{Evaluation of other possible values of loss weight $\lambda$ on BiC.}
    \label{tab: ablation study on lambda bic}
\end{table}
\newpage

\begin{table}[ht]
    \centering
    
    \begin{tabular}{cccccc}
    \toprule
    Method                       & $\lambda_{tcon}$ & $\lambda_{ccon}$ & $\lambda_{ccd}$ & $AA\uparrow$                        & $FM\downarrow$                        \\ \midrule
    \multirow{5}{*}{LwF DisCo-I} & 0.5              & 0.5              & -             & \textbf{56.42}            & 36.52                     \\
                                 & 1.0              & 0.5              & -             & 54.29                     & \textbf{34.15}            \\
                                 & 0.5              & 1.0              & -             & 55.16                     & 35.98                     \\
                                 & 1.0              & 1.0              & -             & 54.60                     & 34.76                     \\ \bottomrule
    \end{tabular}
    \caption{Evaluation of other possible values of loss weight $\lambda$ on LwF.}
    \label{tab: ablation study on lambda LwF}
\end{table}

\begin{table}[ht]
    \centering
    
    \begin{tabular}{cccccc}
    \toprule
    Method                       & $\lambda_{tcon}$ & $\lambda_{ccon}$ & $\lambda_{ccd}$ & $AA\uparrow$                        & $FM\downarrow$                        \\ \midrule
    \multirow{5}{*}{DER DisCo-I} & 0.5              & 0.5              & 1.0             & \textbf{64.87}            & 36.41                     \\
                                 & 1.0              & 0.5              & 1.0             & 60.04                     & 34.83                     \\
                                 & 0.5              & 1.0              & 1.0             & 64.26                     & 37.11                     \\
                                 & 1.0              & 1.0              & 1.0             & 62.34                     & \textbf{34.57}            \\
                                 & 1.0              & 1.0              & 0.5             & \multicolumn{1}{l}{59.26} & \multicolumn{1}{l}{36.98} \\ \bottomrule
    \end{tabular}
    \caption{Evaluation of other possible values of loss weight $\lambda$ on DER.}
    \label{tab: ablation study on lambda DER}
\end{table}

\begin{table}[hb]
    \centering
    
    \begin{tabular}{cccccc}
    \toprule
    Method                       & $\lambda_{tcon}$ & $\lambda_{ccon}$ & $\lambda_{ccd}$ & $AA\uparrow$                        & $FM\downarrow$                       \\ \midrule
    \multirow{5}{*}{L2P DisCo-T} & 0.5              & 0.5              & -             & 83.12                     & 6.80                     \\
                                 & 1.0              & 0.5              & -             & 82.39                     & \textbf{6.65}            \\
                                 & 0.5              & 1.0              & -             & 83.01                     & 6.83                     \\
                                 & 1.0              & 1.0              & -             & \textbf{83.65}            & 6.76                     \\
                                 \bottomrule
    \end{tabular}
    \caption{Evaluation of other possible values of loss weight $\lambda$ on L2P.}
    \label{tab: ablation study on lambda L2P}
\end{table}

\section{Limitations and Future Work}
Our work draws inspiration from the counter-intuitive phenomenon that CIL models demonstrate lower forgetting rate when introducing domain shifts to each task. We use contrastive learning to implicitly simulate this process at the feature level instead of the input level, and this may be not exactly identical to the phenomenon we discover. In the future, we may explore how to use domain shifts explicitly at the input level. For example, we can incorporate some generative networks in our model to insert domain shift into each task. Moreover, at inference time, we need to find a good way to decide which task this sample belongs to, so that it will be easier to decide which domain shift(or which condition) to give to the generative networks.

\end{document}